\documentclass[12pt, a4paper]{article}
\usepackage[utf8]{inputenc}
\usepackage{amsmath, amssymb}
\usepackage{graphicx}
\usepackage{booktabs}
\usepackage{geometry}
\usepackage[authoryear]{natbib}
\usepackage{hyperref}
\usepackage{authblk}
\usepackage{float}     
\usepackage{array}
\usepackage{xcolor}
\usepackage{lmodern}
\usepackage{textcomp}
\usepackage{microtype}

\usepackage{graphicx}
\graphicspath{{./fig/}}

\makeatletter
\def\@maketitle{%
  \newpage
  \null
  \vskip 2em%
  \begin{flushleft}
  \let \footnote \thanks
    {\LARGE \bfseries \@title \par}
    \vskip 1.5em%
    {\large
      \lineskip .5em%
      \@author \par}%
    \vskip 1em%
    {\large \@date}%
  \end{flushleft}%
  \par
  \vskip 1.5em}
\makeatother

\title{cGAP: Generalized Association Plots with HOMALS-Guided Heatmaps for Visualization of High-Dimensional Categorical Data}
\author[1,2]{Chun-houh Chen\thanks{Corresponding author: cchen@stat.sinica.edu.tw}}
\author[3]{Shun-Chuan Chang}
\author[4]{Chiun-How Kao}
\author[1]{Yi-Ju Lee}
\author[2]{Shang-Ying Shiu}
\author[5]{Yin-Jing Tien}
\author[6]{ShengLi Tzeng}
\author[7]{Han-Ming Wu}

\affil[1]{Institute of Statistical Science, Academia Sinica, Taipei, Taiwan}
\affil[2]{Department of Statistics, National Taipei University, New Taipei City, Taiwan}
\affil[3]{Holistic Education Center, Mackay Medical University, New Taipei City, Taiwan}
\affil[4]{Department of Statistics and Data Science, Tamkang University, New Taipei City, Taiwan}
\affil[5]{Institute for Information Industry, Taipei, Taiwan}
\affil[6]{Department of Applied Mathematics and Graduate Institute of Statistics, National Chung Hsing University, Taiwan}
\affil[7]{Department of Statistics, National Chengchi University, Taipei, Taiwan}

\date{}
\begin{document}

\maketitle

\begin{abstract}
High-dimensional categorical data arise in genetics, biomedicine, and the social sciences, yet visualization tools for such data remain far less developed than those for continuous variables. Existing methods either scale poorly, rely heavily on low-dimensional displays detached from the original data matrix, or prioritize predictive accuracy over interpretability. To address this gap, we introduce categorical Generalized Association Plots (cGAP), a visualization framework for nominal, ordinal, and binary data that preserves the original data matrix while augmenting it with interpretable geometric structure. cGAP uses Homogeneity Analysis (HOMALS) to embed subjects and category levels in a three-dimensional Euclidean space and maps the embedding to red-green-blue coordinates so that similar patterns receive similar colors. The framework integrates three coordinated views: a HOMALS-guided heatmap of the raw data matrix, a subject proximity matrix, and a variable proximity matrix. Seriation algorithms are then used to reorder rows and columns to reveal coherent clusters, outliers, and local-to-global structure. We also derive barycentric traceability, projection-distortion, and contrast-preservation properties that clarify how embedding geometry is transferred to the display. We demonstrate the versatility of cGAP through applications to student-animal classification data, mammalian dentition profiles, mushroom records from the UCI Machine Learning Repository, and the Clusters of Orthologous Genes database. These examples show that cGAP supports transparent exploratory analysis by maintaining traceability between derived visual structure and the original categorical observations. cGAP provides a full-matrix, heatmap-based visualization environment for investigating complex categorical datasets across scientific domains.

\end{abstract}

\section{Introduction}
High-dimensional categorical data arise in many domains, including biology, medicine, text analysis, and the social sciences. In these settings, researchers often seek visual representations that reveal clusters, outliers, and association structure while preserving a clear connection to the original observations. However, visualization tools for multivariate categorical data remain less mature than those for continuous data, especially when the number of variables or categories is large.

Existing approaches address this problem from different directions. Classical displays such as bar charts, pie charts, bubble plots, and mosaic plots are useful for low-dimensional categorical data~\citep{hartigan1981mosaics,friendly1994mosaic,friendly1999extending}, but they become difficult to interpret as dimensionality increases.  Embedding-based methods, including homogeneity analysis (HOMALS), multiple correspondence analysis (MCA), and dual scaling, provide low-dimensional geometric representations of categorical structure~\citep{gifi1990nonlinear,michailidis1998gifi,greenacre1984theory,benzecri1973analyse,nishisato1984dual}. These optimal scaling concepts have also been adapted for axis-based displays; for example, the textile plot \citep{kumasaka2008high} extends parallel coordinates by jointly optimizing axis scaling and ordering to align subject trajectories, yielding coordinate representations closely tied to homogeneity analysis. 
While these methods are valuable for exploratory analysis, the resulting low-dimensional or line-based displays do not by themselves preserve the original data matrix as a clean, scalable primary visual object. In parallel, matrix-based visualization methods such as data images and generalized association plots (GAP) preserve observed values and can reveal large-scale structure effectively~\citep{wegman1990hyperdimensional,minnotte1998data,chen2002generalized,chen2004matrix,tien2008methods,wu2010gap}. Yet categorical data pose an additional challenge: unlike continuous data, their values do not naturally share a common quantitative color scale across variables.

This paper introduces categorical Generalized Association Plots (cGAP), a visualization framework for nominal, ordinal, and binary data that combines full-matrix display with interpretable low-dimensional embedding. cGAP uses HOMALS to map subjects and category levels into a three-dimensional Euclidean space, then uses the resulting coordinates to construct color encodings and proximity views for the original matrix. The framework displays three coordinated views: a HOMALS-guided heatmap of the raw data matrix, a subject proximity matrix, and a variable proximity matrix. Through seriation, these views reveal local and global structure while maintaining traceability to the original categorical observations.

Our main contributions are as follows. First, we propose a full-matrix visualization framework for high-dimensional categorical data that preserves the original observation table as a central analytic view. Second, we introduce an interpretable color-construction strategy based on HOMALS embeddings, allowing similar categorical patterns to be represented by similar colors. Third, we integrate a raw-data heatmap with subject-proximity and variable-proximity views in a coordinated display that supports the discovery of clusters, outliers, and multilevel association structure. Fourth, we derive traceability, projection-distortion, and contrast-preservation properties that clarify how embedding geometry is transferred to the display. Fifth, we demonstrate the applicability of cGAP across nominal, ordinal, and binary datasets from multiple scientific domains.

\begin{figure}[htbp]
	\centering
	\includegraphics[width=0.6\textwidth]{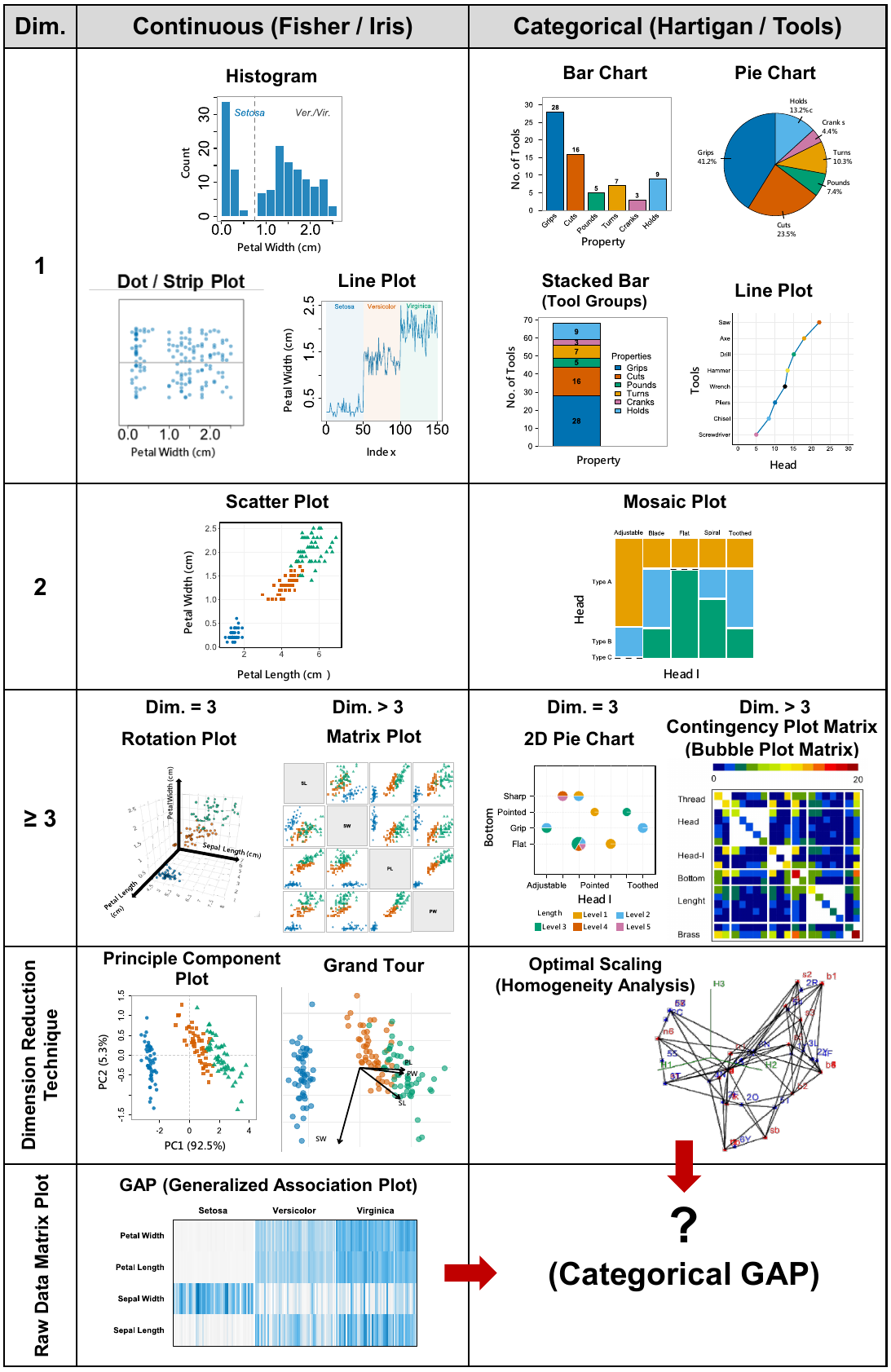} 
	\caption{Graphical tools for continuous data and their categorical counterparts across increasing dimensionality.}
	\label{fig:figure1}
\end{figure}

Figure \ref{fig:figure1} summarizes graphical tools for continuous data and their categorical counterparts across increasing dimensionality. Within this landscape, cGAP is intended as a high-dimensional categorical analogue of matrix-based data-image visualization, while also incorporating embedding-based information to improve interpretability.

The remainder of the paper is organized as follows. Section 2 introduces HOMALS, the measure used to assess the information retained by low-dimensional embedding, and several theoretical properties of the embedding. Section 3 presents the cGAP methodology, including color construction, subject and variable proximity matrices, and contrast enhancement. Section 4 illustrates the method through several applications. Section 5 concludes with discussion and future directions.

\section{HOMALS Embedding of Categorical Data}
cGAP uses homogeneity analysis (HOMALS) to construct a low-dimensional embedding of categorical observations that supports color encoding and proximity-based views. This section summarizes the components of HOMALS that are needed for the cGAP visualization pipeline.

\subsection{Homogeneity Analysis (HOMALS)}
Consider $J$ categorical variables observed on $N$ subjects. Let variable $Z_j$ have $c_j$ categories, and let $G_j$ denote the corresponding $N \times c_j$ indicator matrix, where $G_j(i,t)=1$ if subject $i$ belongs to category $t$ for variable $j$, and 0 otherwise. HOMALS seeks a joint $p$-dimensional Euclidean representation for subjects and category levels. We denote the subject coordinates by $X \in \mathbb{R}^{N \times p}$ and the category coordinates for variable $j$ by $Y_j \in \mathbb{R}^{c_j \times p}$.

The goal is to place subjects near the categories they select and categories near the subjects that select them. This is achieved by minimizing the following average reconstruction loss, where $\mathrm{tr}(\cdot)$ denotes the matrix trace:

\begin{equation}
\sigma(X; Y_1, \dots, Y_J) = J^{-1} \sum_{j=1}^J \text{tr}\left( (X - G_j Y_j)^t (X - G_j Y_j) \right).
\end{equation}

Without additional constraints, the trivial solution $X=0$ and $Y_j=0$ would minimize the objective. We therefore impose centering and normalization constraints so that the embedding is nondegenerate and the dimensions are comparable, using the $N$-vector of ones $u_N$ and the $p \times p$ identity matrix $I_p$:

\begin{equation}
u_N^t X = 0, \quad\mbox{and}\quad
X^t X = N I_p.
\end{equation}

These constraints center the subject configuration at the origin and normalize the embedding. The optimization can be solved by alternating least squares~\citep{deleeuw1967additive}. Given $X$, the category coordinates are updated by least squares:

\begin{equation}
\hat{Y}_j = (G_j^t G_j)^{-1} G_j^t X.
\end{equation}
Given the category coordinates, the subject coordinates are updated as the average of the quantified categories associated with each subject:
\begin{equation}
\hat{X} = J^{-1} \sum_{j=1}^J G_j \hat{Y}_j.
\end{equation}
$X$ is then re-centered and re-scaled until convergence. The resulting embedding has two properties that are especially useful for cGAP. First, category points lie at the centroids of the subjects assigned to them. Second, subjects with identical response patterns receive identical coordinates. These properties make HOMALS a natural bridge between the original categorical matrix and the Euclidean structure later used for color construction and proximity visualization.

\subsection{Information Retention in the Embedding}
Because cGAP maps the HOMALS solution to three color channels, the quality of a low-dimensional embedding matters directly for visualization. The retained structure can be summarized through the optimized loss, which can be written as:

\begin{equation}
\sigma(X; Y_1, \dots, Y_J) = J^{-1} \sum_{j=1}^J \text{tr}\left( (X - G_j Y_j)^t (X - G_j Y_j) \right) = Np - \sum_{j=1}^J \sum_{s=1}^p \eta_{js}^2
\end{equation}
with $\eta_{js}^2$ defined as, where $Y_{j,\cdot s}$ denotes the $s$th column of $Y_j$:

\begin{equation}
\eta_{js}^2 \equiv Y_{j, \cdot s}^t G_j^t G_j Y_{j, \cdot s} / N.
\end{equation}
Here, $\eta_{js}^2$ measures how strongly dimension $s$ discriminates among the categories of variable $j$. The average discrimination power of dimension $s$ across variables is defined as:

\begin{equation}
\gamma_s = \sum_{j=1}^J \eta_{js}^2 / J.
\end{equation}
The associated eigenvalues $\lambda_s = J \gamma_s$ rank the embedding dimensions by the amount of categorical structure they capture. When $p$ is chosen smaller than the full embedding rank $\nu$ of the HOMALS solution, HOMALS acts as a dimension-reduction step. We therefore evaluate the proportion of retained variation using:

\begin{equation}
\gamma_p^* = \frac{\sum_{s=1}^p \gamma_s}{\sum_{s=1}^\nu \gamma_s}.
\end{equation}
This quantity gives the proportion of attainable variation retained by the $p$-dimensional embedding. In cGAP, it provides a practical check on whether a three-dimensional embedding is adequate for downstream color encoding and structural visualization.

\subsection{Theoretical Properties of the Embedding}
The HOMALS solution also yields several structural properties that are directly relevant to the interpretability of cGAP. Let $c_j(i)$ denote the category selected by subject $i$ for variable $j$, let $\hat{x}_i$ denote the $i$th row of $\hat{X}$, and let $\hat{y}_{j,t}$ denote the $t$th row of $\hat{Y}_j$, that is, the fitted coordinate vector of category $t$ for variable $j$.

\paragraph{Proposition 1 (Barycentric Traceability)} For each subject $i$, the HOMALS coordinate is the barycenter of the category coordinates selected by that subject:
\begin{equation}
\hat{x}_i = \frac{1}{J} \sum_{j=1}^J \hat{y}_{j,c_j(i)}.
\end{equation}
The result follows immediately by taking the $i$th row of the update equation $\hat{X} = J^{-1} \sum_{j=1}^J G_j \hat{Y}_j$. Since each row of $G_j$ contains exactly one nonzero entry, the $i$th row of $G_j \hat{Y}_j$ is the coordinate of the category selected by subject $i$ for variable $j$.

This proposition formalizes the traceability of cGAP: each subject point is an average of the categories that define its response profile. Because the RGB relocation used in Section 3.1 is affine, the same identity also holds for displayed colors.

\paragraph{Corollary 1 (Affine Color Traceability)} For a three-dimensional display, let $m$ denote the maximum absolute coordinate over all fitted subject and category points, and let $\mathbf{1}_3$ denote the three-vector of ones. Define the RGB relocation map $T(z)=z/(2m)+0.5\mathbf{1}_3$ coordinatewise. Then

\begin{equation}
T(\hat{x}_i) = \frac{1}{J} \sum_{j=1}^J T(\hat{y}_{j,c_j(i)}).
\end{equation}

Hence the displayed color of a subject is the affine average of the colors of its selected categories. This identity justifies the profile-color interpretation used throughout cGAP.

\paragraph{Proposition 2 (Projection-Distortion Decomposition)} Let $\nu$ denote the full embedding rank of the HOMALS solution. Let $\hat{x}_i^{(\nu)} \in \mathbb{R}^{\nu}$ denote the full-rank HOMALS coordinate of subject $i$, and let $\hat{x}_i^{(p)}$ denote its first $p$ coordinates. Then for any two subjects $i$ and $i'$,
\begin{equation}
\left\| \hat{x}_i^{(\nu)} - \hat{x}_{i'}^{(\nu)} \right\|^2
=
\left\| \hat{x}_i^{(p)} - \hat{x}_{i'}^{(p)} \right\|^2
+ \sum_{s=p+1}^{\nu} \left( \hat{x}_{is} - \hat{x}_{i's} \right)^2.
\end{equation}
Expand the squared Euclidean norm componentwise, where $\hat{x}_{is}$ denotes the $s$th component of $\hat{x}_i^{(\nu)}$, and separate the first $p$ coordinates from the discarded coordinates.

Therefore, the $p$-dimensional display can only underestimate full-space pairwise distances, and the discrepancy is determined exactly by the discarded coordinates. The same decomposition applies to category coordinates. This result complements the retained-variation criterion $\gamma_p^*$ by giving a geometric interpretation of the distortion induced by restricting the display to three dimensions.

\section{Constructing cGAP Displays}

This section describes how cGAP transforms a categorical data table into three coordinated visual representations: a HOMALS-guided heatmap of the raw data matrix, a subject proximity matrix, and a variable proximity matrix. We use the animal-grouping data of~\citet{nishisato2007multidimensional} as a running example. In this dataset, fifteen students grouped thirty-five animals according to perceived similarity, providing a compact illustration of nominal categorical structure.

\begin{table}[h]
\centering
\caption{Animal-grouping data from fifteen students. Each entry records the group assigned by a student to an animal. The numbers of groups used by the fifteen students were 8, 3, 9, 9 (with no group 9 for Student 4), 7 (Student 5 used groups 0$-$6), 5, 7, 5, 5, 5, 6, 4, 8, 7, and 7 (with no group 7 for Student 15), producing ninety-five student-group categories in total.}
\label{tab:table1}
\resizebox{0.6\textwidth}{!}{
\begin{tabular}{lccccccccccccccc}
 & \multicolumn{15}{c}{Students} \\
\cmidrule(lr){2-16}
Animals & 1 & 2 & 3 & 4 & 5 & 6 & 7 & 8 & 9 & 10 & 11 & 12 & 13 & 14 & 15 \\
\midrule
Dog & 1 & 1 & 1 & 1 & 1 & 1 & 1 & 1 & 1 & 1 & 1 & 1 & 1 & 1 & 1 \\
Alligator & 2 & 2 & 2 & 2 & 2 & 2 & 3 & 3 & 2 & 2 & 2 & 2 & 3 & 2 & \textbf{8} \\
Chimpanzee & 3 & 1 & 3 & 3 & 3 & 1 & 4 & 3 & 3 & 3 & 3 & 2 & 3 & 2 & 3 \\
Cow & 4 & 1 & 4 & 4 & 4 & 2 & 3 & 4 & 1 & 2 & 1 & 1 & 1 & 3 & 4 \\
Crow & 5 & \textbf{3} & 5 & 5 & 5 & 4 & 2 & \textbf{5} & 4 & 4 & 4 & \textbf{4} & 4 & 6 & 5 \\
Pigeon & 5 & 3 & 5 & 5 & 5 & 4 & 2 & 5 & 4 & 4 & 4 & 1 & 4 & 6 & 5 \\
Cheetah & 6 & 1 & 1 & 6 & 1 & 2 & 3 & 4 & \textbf{5} & 1 & 3 & 1 & 2 & 4 & 2 \\
Chicken & 5 & 3 & 5 & \textbf{10} & 5 & 4 & 6 & 5 & 4 & 4 & 1 & 4 & 4 & 6 & 5 \\
Bear & 1 & 1 & 6 & 4 & 2 & 2 & 3 & 3 & 5 & \textbf{5} & 5 & 1 & 1 & 4 & 2 \\
Cat & 1 & 1 & 1 & 6 & 1 & 2 & 1 & 1 & 1 & 1 & 1 & 1 & 2 & 1 & 1 \\
Rabbit & 1 & 1 & 7 & 4 & 1 & \textbf{5} & \textbf{7} & 1 & 1 & 1 & 1 & 1 & 5 & 1 & 1 \\
Frog & 2 & 2 & 8 & 9 & \textbf{6} & 3 & 5 & 2 & 2 & 2 & \textbf{6} & 3 & 6 & \textbf{7} & 6 \\
Goat & 6 & 1 & 4 & 4 & 4 & 5 & 7 & 4 & 1 & 1 & 1 & 1 & 5 & 1 & 1 \\
Tiger & 3 & 1 & 1 & 6 & 1 & 2 & 3 & 4 & 5 & 1 & 2 & 1 & 2 & 4 & 2 \\
Rhinoceros & 7 & 1 & 3 & 4 & 2 & 5 & 3 & 4 & 5 & 2 & 5 & 1 & 7 & 5 & 2 \\
Giraffe & 4 & 1 & \textbf{9} & 7 & 4 & 5 & 3 & 4 & 5 & 3 & 5 & 1 & 7 & 5 & 4 \\
Duck & 5 & 3 & 5 & 10 & 5 & 4 & 2 & 5 & 4 & 4 & 4 & 4 & 4 & 6 & 5 \\
Sparrow & 5 & 3 & 5 & 5 & 5 & 4 & 2 & 5 & 4 & 3 & 4 & 4 & 4 & 6 & 5 \\
Hippopotamus & 7 & 1 & 2 & 4 & 2 & 5 & 3 & 2 & 5 & 2 & 5 & 1 & 7 & 5 & 2 \\
Monkey & 1 & 1 & 3 & 3 & 3 & 1 & 4 & 3 & 3 & 3 & 3 & 2 & 3 & 2 & 3 \\
Turkey & 5 & 3 & 5 & 10 & 5 & 4 & 6 & 5 & 4 & 4 & 1 & 4 & 4 & 6 & 5 \\
Pig & 4 & 1 & 4 & 4 & 4 & 2 & 7 & 4 & 1 & 1 & 1 & 1 & 5 & 3 & 1 \\
Crane & 5 & 3 & 5 & 5 & 5 & 4 & 2 & 5 & 4 & 4 & 4 & 4 & 4 & 6 & 5 \\
Leopard & 4 & 1 & 1 & 6 & 1 & 2 & 3 & 4 & 5 & 1 & 2 & 1 & 2 & 4 & 2 \\
Ostrich & 6 & 3 & 5 & 10 & 5 & 4 & 3 & 5 & 4 & 4 & 5 & 4 & 4 & 5 & 4 \\
Lizard & \textbf{8} & 2 & 8 & 9 & 6 & 3 & 5 & 2 & 2 & 2 & 6 & 3 & \textbf{8} & 7 & 6 \\
Horse & 1 & 1 & 4 & 4 & 4 & 1 & 3 & 4 & 1 & 1 & 1 & 1 & 5 & 3 & 4 \\
Raccoon & 1 & 1 & 1 & 1 & 0 & 2 & 7 & 1 & 1 & 1 & 2 & 1 & 1 & 1 & 1 \\
Tortoise & 2 & 2 & 8 & 2 & 6 & 3 & 5 & 2 & 2 & 5 & 6 & 3 & 6 & 7 & 6 \\
Snake & 2 & 2 & 8 & 2 & 2 & 3 & 5 & 2 & 2 & 2 & 6 & 3 & 8 & 7 & 6 \\
Lion & 3 & 1 & 1 & 6 & 2 & 2 & 3 & 4 & 5 & 1 & 2 & 1 & 7 & 4 & 2 \\
Elephant & 7 & 1 & 2 & 7 & 4 & 5 & 3 & 4 & 5 & 5 & 2 & 1 & 7 & 5 & 2 \\
Camel & 7 & 1 & 4 & 7 & 4 & 5 & 3 & 4 & 5 & 3 & 5 & 1 & 7 & 5 & 4 \\
Hawk & 5 & 3 & 5 & 5 & 5 & 4 & 2 & 5 & 4 & 4 & 2 & 4 & 4 & 6 & 5 \\
Fox & 1 & 1 & 1 & 6 & 1 & 2 & 7 & 1 & 1 & 1 & 2 & 1 & 1 & 1 & 1 \\
\bottomrule
\end{tabular}%
}
\end{table}

\subsection{Color Encoding from the HOMALS Embedding}

The main challenge in extending matrix-based visualization from continuous to categorical data is color construction. For continuous variables, a common numerical scale permits a single ordered color spectrum. For multivariate categorical data, category labels are variable-specific and do not share a global ordering. cGAP addresses this by deriving colors from the geometry of the HOMALS embedding rather than from the labels themselves. This choice is designed to satisfy the relativity principle of statistical graphics~\citep{chen2002generalized}: subjects and categories that are close in the embedding should receive similar colors, whereas distant configurations should receive more distinct colors.

We use a three-dimensional HOMALS embedding because it can be mapped directly to the three channels of a standard display device. RGB is adopted here as a display-native coordinate system, not as a claim that it is perceptually optimal among all color spaces. Its practical advantage is that each embedding dimension can be assigned to one channel, allowing the geometry of the embedding to be transferred transparently to color. The analytic content of cGAP therefore lies in relative color similarity and contrast, rather than in any fixed semantic meaning of red, green, or blue individually.

As with other low-dimensional embeddings, HOMALS axes are unique up to sign changes and axis permutation. In cGAP, this affects which absolute hues appear in a figure, but it does not alter the underlying proximity structure. Pairwise distances, cluster relationships, and block patterns in the sorted matrices remain unchanged. Throughout this paper, we order the embedding axes by decreasing retained variation (H1, H2, H3) and map them to the R, G, and B channels, respectively. Under this convention, color should be interpreted relationally: similar colors indicate similar embedded configurations, even though a particular hue does not by itself carry a fixed substantive meaning.

To map the embedding into display colors, we first relocate subjects and categories into the unit cube $[0, 1]^3$. Let $x \in [0,1]^{N \times 3}$ and $y_j \in [0,1]^{c_j \times 3}$ denote the relocated subject and category coordinate matrices, respectively. For subject $i$, category $c$ of variable $j$, and RGB channel $r \in \{1,2,3\}$, define
\begin{equation}
x(i, r) = \frac{\hat{X}(i, r)}{2m} + 0.5
\end{equation}

\begin{equation}
y_j(c, r) = \frac{\hat{Y}_j(c, r)}{2m} + 0.5,
\end{equation}
where $m$ is the maximum absolute coordinate over all fitted subject and category points before relocation. The first, second, and third coordinates then define the red, green, and blue intensities. This linear rescaling preserves the relative arrangement of points while making them displayable as colors. Points near the center of the configuration are mapped to grayish tones, whereas more extreme configurations move toward the cube faces and corners. Edges connecting subjects to categories inherit the category color, visually linking the joint embedding to the original matrix entries.

\begin{figure}[t]
	\centering
	\includegraphics[scale=0.3,keepaspectratio]{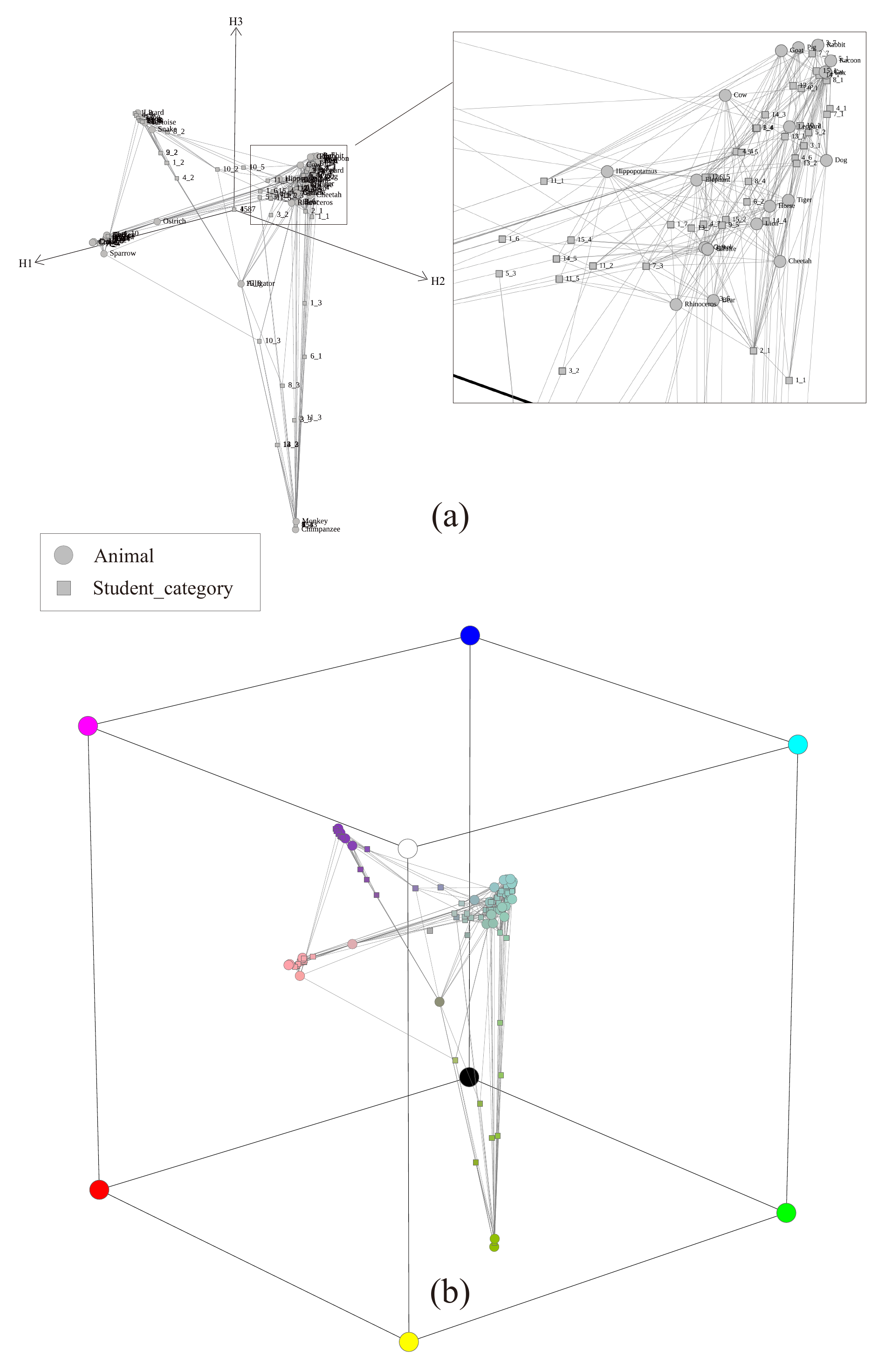} 
	\caption{(a) The three-dimensional HOMALS embedding, where H1, H2, and H3 denote the first three dimensions. (b) The same configuration after linear relocation into the unit RGB cube. Edges connect subjects to the categories they select.}
	\label{fig:figure2}
\end{figure}

Figure \ref{fig:figure2}(a) shows the fitted HOMALS embedding, and Figure \ref{fig:figure2}(b) shows the same configuration after relocation into the RGB unit cube. The resulting category colors are displayed in Figure \ref{fig:figure3}(a) and are then used to construct the raw-data heatmap in Figure \ref{fig:figure3}(b). Representative colors for animals are obtained by averaging the colors of the student-group categories associated with each animal (Figure \ref{fig:figure3}(c)), consistent with the dual representation property of HOMALS before optional contrast enhancement. Because cGAP is a color-centered visualization, interpretation should not rely on hue alone. The method is intended to be read jointly through the sorted raw-data heatmap, the subject proximity matrix, the variable proximity matrix, and the associated dendrograms. A more comprehensive study of perceptually optimized palettes and color-vision-deficiency-safe encodings remains an important direction for future work.

\subsection{Subject Proximity}
For subjects, cGAP uses Euclidean distance in the three-dimensional HOMALS embedding as the proximity measure. Figure \ref{fig:figure3}(d) shows the resulting between-animal proximity matrix for the data in Table \ref{tab:table1}.

\subsection{Variable Proximity}
For categorical variables, cGAP defines an embedding-based dissimilarity in terms of the HOMALS category coordinates. Let $Z_k$ be a categorical variable with $c_k$ categories, let $\omega_i$ denote subject $i$, and let $Z_k(\omega_i)$ denote the response of subject $i$ on variable $Z_k$. Define $n_t^k = \#\{\omega_i : Z_k(\omega_i)=t\}$ as the count for category $t$ of variable $Z_k$, and $n_{ts}^{kl} = \#\{\omega_i : Z_k(\omega_i)=t \ \& \ Z_l(\omega_i)=s\}$ as the joint count for categories $t$ and $s$ of variables $Z_k$ and $Z_l$. Since $\sum_{t=1}^{c_k} n_t^k = \sum_{t=1}^{c_k} \sum_{s=1}^{c_l} n_{ts}^{kl} = N$, we define the dissimilarity between $Z_k$ and $Z_l$ as:

\begin{equation}
d(Z_k, Z_l) = \sum_{i=1}^N D(Z_k(\omega_i), Z_l(\omega_i)) = \sum_{t=1}^{c_k} \sum_{s=1}^{c_l} n_{ts}^{kl} D(t,s),
\end{equation}
where $D(t,s)$ denotes the Euclidean distance between category $t$ of $Z_k$ and category $s$ of $Z_l$ in the $p$-dimensional HOMALS embedding. This function $d$ satisfies the four metric conditions:
\begin{itemize}
	\item \textbf{Non-negativity:} Since $D(t, s) \geq 0$, it follows that $d(Z_k, Z_l) \geq 0$.
	
	\item \textbf{Identity of Indiscernibles:} If $Z_k = Z_l$ (i.e., there exists a function $g$ such that $g(Z_k(\omega_i)) = Z_l(\omega_i), \forall \omega_i$), then the HOMALS solutions for $Z_k$ and $Z_l$ are identical and hence $d(Z_k, Z_l) = 0$; conversely, if $Z_k \neq Z_l$ there exists at least one $\omega_i$ such that $D(Z_k(\omega_i), Z_l(\omega_i)) > 0$, so $d(Z_k, Z_l) > 0$.
	
	\item \textbf{Symmetry:} The Euclidean distance is symmetric, so $d(Z_k, Z_l) = d(Z_l, Z_k)$.
	
	\item \textbf{Triangle Inequality:} Since $D(Z_k(\omega_i), Z_l(\omega_i)) \leq D(Z_k(\omega_i), Z_m(\omega_i)) + D(Z_m(\omega_i), Z_l(\omega_i))$ for any third variable $Z_m$ and all $\omega_i$, it follows that $d(Z_k, Z_l) \leq d(Z_k, Z_m) + d(Z_m, Z_l)$.
\end{itemize}

We interpret $d(Z_k, Z_l)$ as a weighted embedding-based dissimilarity, rather than as a strict metric on the original variable space. It is nonnegative and symmetric, and because it aggregates Euclidean distances across subjects it also inherits the triangle inequality. A zero value indicates that the two variables are indistinguishable under this induced comparison for the observed data; it should not be interpreted as semantic identity of the original variables. This level of structure is sufficient for cGAP, whose goal is to organize variables by empirical proximity in the embedding for visualization and seriation.

\begin{figure}[h]
    \centering
    \includegraphics[width=0.6\textwidth]{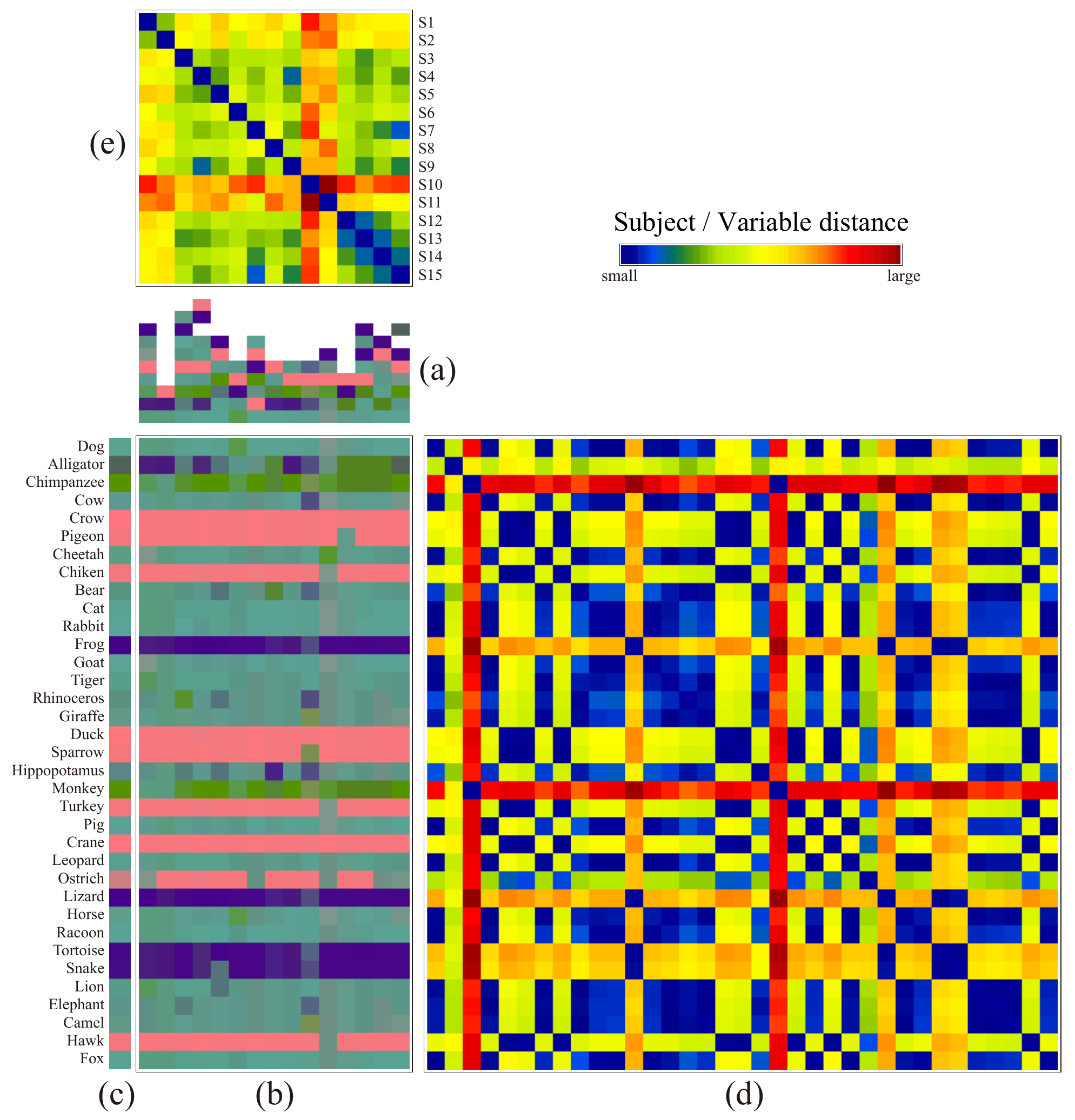} 
    \caption{The unsorted cGAP display for the animal-grouping data in Table \ref{tab:table1} (35 animal samples $\times$ 15 student variables): (a) category color map, (b) the raw-data heatmap, (c) animal profile colors, (d) the between-animal Euclidean distance map, and (e) the between-student weighted distance map.}
    \label{fig:figure3}
\end{figure}

Figure \ref{fig:figure3}(e) displays the resulting between-student proximity matrix for the data in Table \ref{tab:table1}.

\subsection{Integrated Display and Seriation}

Having defined the color map and the two proximity matrices, cGAP assembles them into a coordinated display. Figure \ref{fig:figure3} presents the animal-student data in its original order. Even before reordering, the combined use of the raw-data heatmap and proximity views reveals several recognizable features, including broad animal groupings, heterogeneous treatment of the alligator, the similarity of chimpanzee and monkey, and unusual response patterns for Students S10 and S11.

The next step is seriation. GAP algorithms reorder the rows and columns of the raw-data heatmap and the two proximity matrices so that similar subjects and variables appear adjacent. In this paper, we use seriation methods built around Rank-Two Ellipse (R2E), Rank-One Tree (R1T), and the hybrid HCT-R2E procedure. R2E emphasizes smooth global trends, hierarchical clustering tree (HCT) seriation stabilizes local neighborhood structure, and HCT-R2E combines these behaviors by flipping internal HCT nodes under guidance from the R2E ordering. We use average linkage for agglomerative HCT and also consider the analogous R1T-R2E construction for divisive ordering.

\begin{figure}[htbp]
    \centering
    \includegraphics[width=0.70\textwidth]{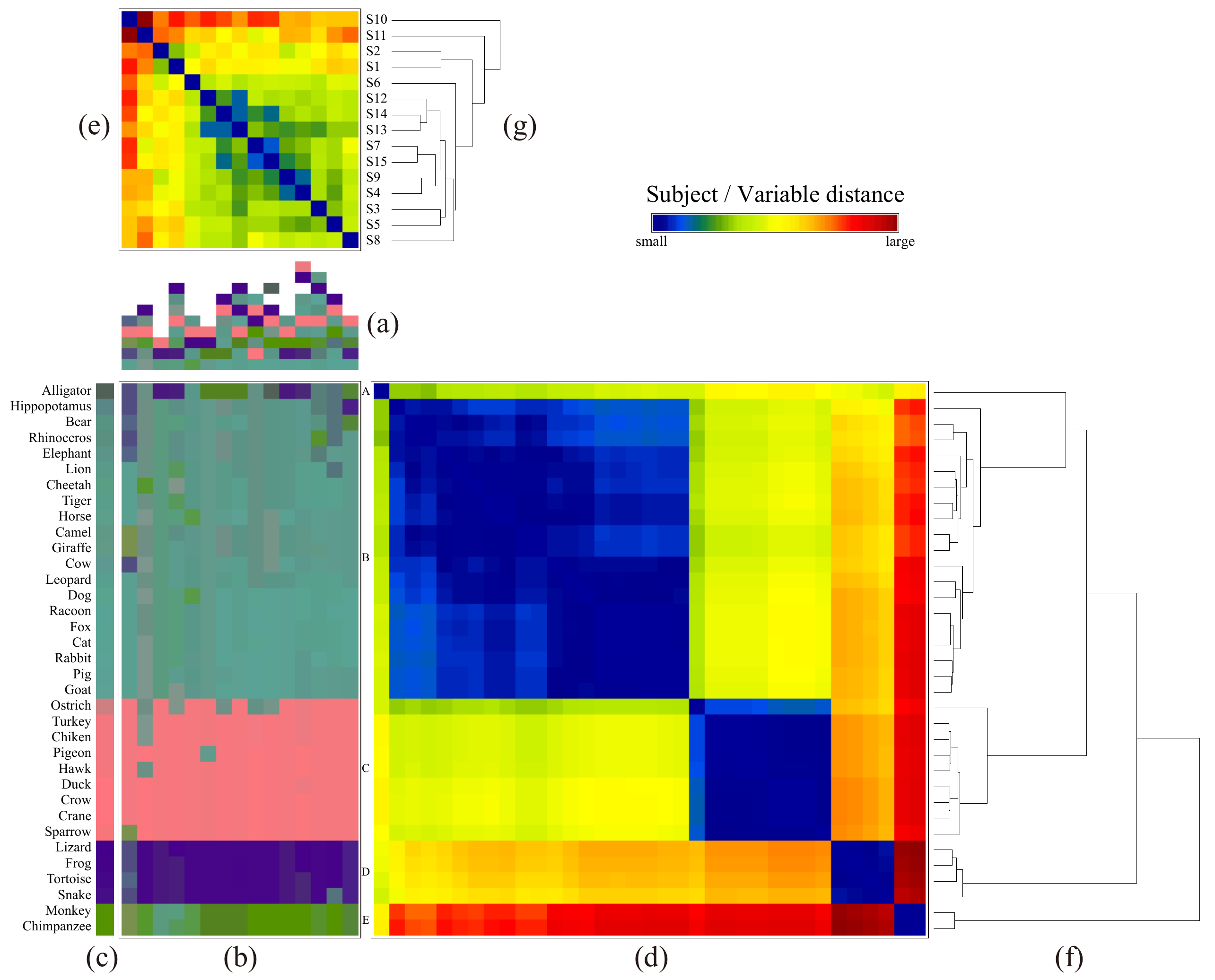} 
    \caption{cGAP with HCT-R2E seriation for the animal-grouping data in Table \ref{tab:table1} (35 animal samples $\times$ 15 student variables): (a) category color map, (b) the sorted raw-data heatmap, (c) animal profile colors, (d) the sorted Euclidean distance map for animals, (e) the sorted weighted distance map for students, (f) the HCT-R2E tree for (d), and (g) the HCT-R2E tree for (e).}
    \label{fig:figure4}
\end{figure}

Figure \ref{fig:figure4} shows the cGAP results after applying the HCT-R2E seriation algorithm for both the animals (samples) and students (variables). The color patterns in the sorted raw-data heatmap (Figure \ref{fig:figure4}(b)) and animal profiles (Figure \ref{fig:figure4}(c)) identify four main perceptual groups: B. non-primate mammals (sea green), C. birds (pink), D. reptiles and frogs (purple), and E. primates (green). The sorted between-animal distance matrix (Figure \ref{fig:figure4}(d)) clearly highlights the block-diagonal patterns of the four animal groups, featuring a stand-alone cluster (A) of alligators at the top-left.

Notably, the alligator elicited diverse classifications, with students assigning it variously to all four possible animal groups or a combination thereof. Figure \ref{fig:figure4}(c) renders the color of the alligator as a blend of purple, sea green, and green. This visual averaging effectively illustrates the principle of relativity in statistical graphics. The ostrich also appears distinct, serving as a bridging animal between the two major clusters of mammals and birds in the proximity matrix (Figure \ref{fig:figure4}(b)), reflecting its categorization with mammals by five students (S1, S7, S11, S14, S15). Meanwhile, the monkey and chimpanzee share nearly identical color profiles and high proximity, indicating they were consistently grouped together by almost all students. Furthermore, the distinct red off-diagonal regions between the primate cluster and other groups in Figure \ref{fig:figure4}(d) visually signify a large distance, confirming their dissimilarity from other animals. The branching patterns of the HCT in Figure \ref{fig:figure4}(d) further delineate the four major animal groups along with the unique relationships of the alligator and ostrich to the other animals.

Fundamentally, Figure \ref{fig:figure4}(a) presents the color representation of the categories, where categories sharing compositional similarity are assigned comparable hues. For instance, in Student S8's classification of 35 animals into five groups, the first, third, and fourth categories, comprising mammals, are rendered in varying shades of sea green. In contrast, the second and fifth categories, which contain non-mammals, are depicted in distinct hues such as purple and pink. Similarly, although Student S5 assigned nine birds to the sixth category and Student S8 assigned them to the fifth, both categories yield identical HOMALS solutions and are consequently displayed in the same pink hue. This confirms that the color representation of categories strictly adheres to the relativity principle.

\subsection{Contrast Enhancement for Outlying Configurations}

Figures \ref{fig:figure3} and \ref{fig:figure4} also illustrate how cGAP exposes atypical samples, variables, and categories. In this dataset, the alligator, the ostrich, and the primate cluster occupy distinctive positions, and Students S10 and S11 show response patterns that differ from those of most other students. Additional local anomalies are also visible, such as the treatment of pigeon by Student S12 and the treatment of sparrow and cow by Student S10. These observations highlight the exploratory role of cGAP: unusual configurations can be identified visually and then investigated substantively in context.

A technical challenge arises when outlying configurations lie far from the data center in HOMALS space. After relocation into the unit cube, such points can pull the remaining configurations toward the center, reducing color contrast for the majority of the data. To mitigate this, we apply a power transformation to the relocated coordinates. Let $x \in [0,1]^{N \times 3}$ and $y_j \in [0,1]^{c_j \times 3}$ denote the relocated subject and category coordinate matrices defined above, and let
\[
z = [x^t\ y_1^t\ \cdots\ y_J^t]^t
\]
be the resulting $M \times 3$ matrix of display points, where $M = N + \sum_{j=1}^J c_j$. For a contrast parameter $q \ge 1$, define the transformed matrix $z^{(q)}$ by
\[
z^{(q)}(k,r) = \frac{z(k,r) - 0.5}{w_k} w_k^{\frac{1}{q}} + 0.5,
\]
for $k \in \{1, 2, \dots, M\}$, $r \in \{1, 2, 3\}$, and $w_k = \max_{1 \le r \le 3}|2z(k,r) - 1|$.

This transformation moves points outward from the center and enhances visual contrast. When $q=1$, the display is unchanged because $z^{(1)} = z$; as $q \to \infty$, points are progressively pushed toward the cube surfaces. In practice, this adjustment improves visual separation while preserving the relational logic of the color encoding.

\paragraph{Proposition 3 (Ray-Preserving Contrast Transform)} Let $\mathbf{1}_3$ be the three-vector of ones and define the centered coordinate $u_k = z(k,:) - 0.5 \mathbf{1}_3$. Then the transformed point satisfies

\begin{equation}
z^{(q)}(k,:) - 0.5 \mathbf{1}_3 = w_k^{\frac{1}{q}-1}\bigl(z(k,:) - 0.5 \mathbf{1}_3\bigr).
\end{equation}
Apply the transformation coordinatewise and factor out the common multiplier $w_k^{1/q-1}$, and the result follows. Because the multiplier is positive, each point remains on the same ray from the center of the RGB cube, its sign pattern relative to the center is unchanged, and any two points on a common ray retain their radial order. Thus the transform changes saturation and visual contrast without altering the directional structure inherited from HOMALS.

Together, Table \ref{tab:table1} and Figures \ref{fig:figure2}$-$\ref{fig:figure4} show how cGAP combines embedding, color encoding, proximity structure, and seriation to reveal multilevel categorical patterns in a single exploratory environment. The next section presents larger examples spanning ordinal, nominal, and binary data.

\section{Applications Across Data Types}
We evaluate cGAP on three datasets chosen to test complementary aspects of categorical visualization: an ordinal biological dataset (mammalian dentition), a nominal benchmark with interpretable class structure (mushrooms), and a large-scale binary genomics dataset (COG profiles). Together, these examples assess how cGAP handles different measurement scales, different levels of structural complexity, and different analysis goals, from compact interpretable clustering problems to large exploratory screening tasks (Table \ref{tab:table2}).

\begin{table}[htbp]
\centering
\caption{Summary of the application examples used to evaluate cGAP.}
\small
\renewcommand{\arraystretch}{1.15}
\setlength{\tabcolsep}{4pt}
\begin{tabular}{@{}
>{\raggedright\arraybackslash}p{0.18\textwidth}
>{\raggedright\arraybackslash}p{0.09\textwidth}
>{\raggedright\arraybackslash}p{0.15\textwidth}
>{\raggedright\arraybackslash}p{0.22\textwidth}
>{\raggedright\arraybackslash}p{0.26\textwidth}@{}}
\toprule
Dataset & Scale & Matrix size & Role in evaluation & What cGAP reveals \\
\midrule
Mammalian dentition & Ordinal & 66 $\times$ 8 & Biological ordinal structure & Anatomical symmetry, functional dentition gradients, and taxonomy-aligned exceptions \\
Mushroom attributes & Nominal & 8,124 $\times$ 22 & Interpretable class-related structure & Locally decisive variables, edibility regimes, and informative missingness \\
COG profiles & Binary & 2,296 $\times$ 5,061 & Large-scale exploratory screening & Phylogenetic blocks, complementary present/absent modules, and multiscale genomic structure \\
\bottomrule
\end{tabular}
\label{tab:table2}
\end{table}

\subsection{Mammalian Dentition: Ordinal Biological Structure}

The mammalian dentition dataset, introduced by~\citet{hartigan1975clustering} and later analyzed with homogeneity analysis by~\citet{michailidis1998gifi}, contains 66 mammals described by eight dentition variables. These variables record the numbers of incisors, canines, premolars, and molars on the upper and lower jaws. Because the variables are ordinal, this example tests whether cGAP can preserve ordered categorical structure while still revealing biologically meaningful groupings. Table \ref{tab:table3} lists the full data and the taxonomic order of each mammal.

\begin{figure}[htbp]
    \centering
    \includegraphics[width=0.70\textwidth]{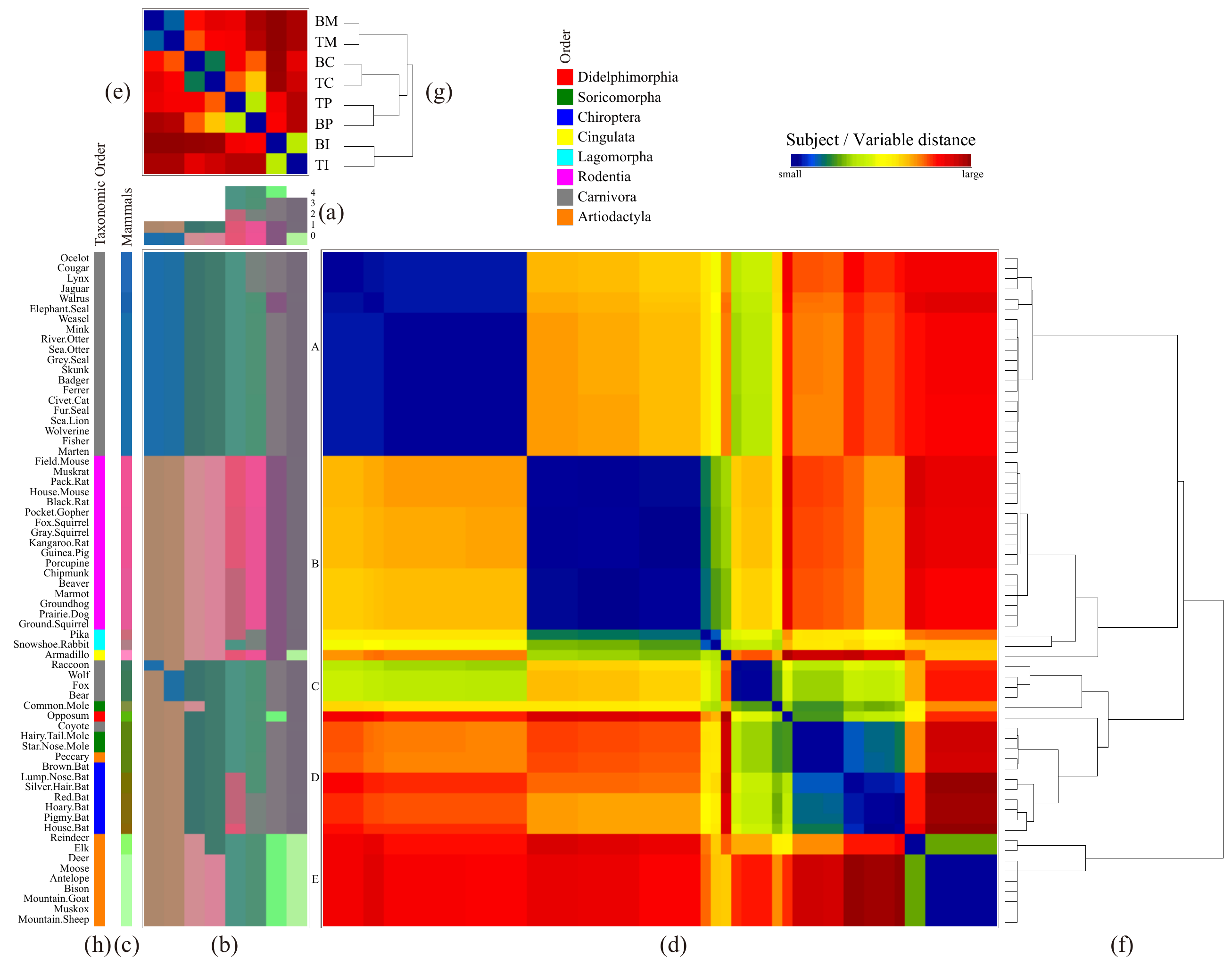} 
    \caption{cGAP analysis of the mammalian dentition data (66 mammals $\times$ 8 ordinal variables): (a) category color map, (b) sorted raw-data heatmap, (c) mammal profile colors, (d) sorted Euclidean distance map for mammals, (e) sorted weighted distance map for dentition variables, (f) HCT-R2E for (d), (g) HCT-R2E for (e), and (h) mammal names with taxonomic-order color legend.}
    \label{fig:figure5_dentition}
\end{figure}

We fit HOMALS with order constraints and obtain a three-dimensional embedding that retains approximately 67\% of the attainable variation. Figure \ref{fig:figure5_dentition} shows the resulting cGAP display after HCT-R2E seriation. This example is useful because the dentition variables have clear biological meaning and the taxonomic labels provide an external reference for interpretation.

The variable proximity view shows that cGAP recovers coherent ordinal structure rather than isolated variable effects. In particular, the top and bottom versions of the same tooth types lie close together, especially for molars and canines. The category colors also reveal an interpretable opposition between these tooth patterns: mammals with fewer molars tend to retain canines, whereas mammals with more molars often do not. This relationship is expressed simultaneously in the category configuration and in the sorted data matrix.

The subject-oriented views further reveal clusters that align well with taxonomy. Carnivora mammals form coherent groups, whereas Rodentia- and Lagomorpha-related mammals occupy a separate region characterized by more molars and fewer canines. Artiodactyla mammals are grouped by the joint absence of incisors and canines. The display also isolates biologically distinctive animals such as the armadillo, whose extreme dentition profile appears as an outlying configuration in both color and proximity space. Overall, Figure \ref{fig:figure5_dentition} shows that cGAP can handle ordered categorical variables without collapsing the data to a single low-dimensional scatterplot, while still linking the observed patterns back to the original table.

A further strength of this example is that cGAP reveals not only taxonomic clustering but also structural regularities that cut across taxonomy. The close placement of upper and lower tooth variables indicates that the display captures paired anatomical organization rather than treating each measurement independently. At the sample level, the sorted heatmap suggests a biologically interpretable gradient from carnivore-like dentition, characterized by retained canines and reduced molar counts, toward herbivore- and rodent-like dentition, characterized by reduced canines and relatively expanded molar structure. The display also preserves informative exceptions within this gradient. For example, the Common Mole appears closer to a carnivore-like profile than its taxonomic label alone would suggest, whereas the armadillo forms an extreme boundary case that anchors one end of the dentition space. This combination of symmetry, gradient structure, and localized exceptions illustrates how cGAP can expose functional morphological relationships in ordinal biological data while keeping every observed tooth pattern visible in the raw-data heatmap.

\begin{table}[t]
\centering
\caption{Dentition data for 66 mammals. Variables are top incisors (TI), bottom incisors (BI), top canines (TC), bottom canines (BC), top premolars (TP), bottom premolars (BP), top molars (TM), bottom molars (BM), and taxonomic order: Didelphimorphia (D), Soricomorpha (S), Chiroptera (Ch), Cingulata (Ci), Lagomorpha (L), Rodentia (R), Carnivora (Ca), and Artiodactyla (A).}
\label{tab:table3}
\resizebox{\textwidth}{!}{%
\begin{tabular}{l cccccccc c | l cccccccc c}
\toprule
\textbf{Mammal} & \textbf{TI} & \textbf{BI} & \textbf{TC} & \textbf{BC} & \textbf{TP} & \textbf{BP} & \textbf{TM} & \textbf{BM} & \textbf{Order} & 
\textbf{Mammal} & \textbf{TI} & \textbf{BI} & \textbf{TC} & \textbf{BC} & \textbf{TP} & \textbf{BP} & \textbf{TM} & \textbf{BM} & \textbf{Order} \\
\midrule
Opossum & 3 & 4 & 1 & 1 & 3 & 3 & 1 & 1 & D & Fox & 3 & 3 & 1 & 1 & 4 & 4 & 0 & 1 & Ca \\
Hairy Tail Mole & 3 & 3 & 1 & 1 & 4 & 4 & 1 & 1 & S & Bear & 3 & 3 & 1 & 1 & 4 & 4 & 0 & 1 & Ca \\
Common Mole & 3 & 2 & 1 & 0 & 3 & 3 & 1 & 1 & S & Civet Cat & 3 & 3 & 1 & 1 & 4 & 4 & 0 & 0 & Ca \\
Star Nose Mole & 3 & 3 & 1 & 1 & 4 & 4 & 1 & 1 & S & Raccoon & 3 & 3 & 1 & 1 & 4 & 4 & 1 & 0 & Ca \\
Brown Bat & 2 & 3 & 1 & 1 & 3 & 3 & 1 & 1 & Ch & Marten & 3 & 3 & 1 & 1 & 4 & 4 & 0 & 0 & Ca \\
Silver Hair Bat & 2 & 3 & 1 & 1 & 2 & 3 & 1 & 1 & Ch & Fisher & 3 & 3 & 1 & 1 & 4 & 4 & 0 & 0 & Ca \\
Pigmy Bat & 2 & 3 & 1 & 1 & 2 & 2 & 1 & 1 & Ch & Weasel & 3 & 3 & 1 & 1 & 3 & 3 & 0 & 0 & Ca \\
House Bat & 2 & 3 & 1 & 1 & 1 & 2 & 1 & 1 & Ch & Mink & 3 & 3 & 1 & 1 & 3 & 3 & 0 & 0 & Ca \\
Red Bat & 1 & 3 & 1 & 1 & 2 & 2 & 1 & 1 & Ch & Ferret & 3 & 3 & 1 & 1 & 3 & 3 & 0 & 0 & Ca \\
Hoary Bat & 1 & 3 & 1 & 1 & 2 & 2 & 1 & 1 & Ch & Wolverine & 3 & 3 & 1 & 1 & 4 & 4 & 0 & 0 & Ca \\
Lump Nose Bat & 2 & 3 & 1 & 1 & 2 & 3 & 1 & 1 & Ch & Badger & 3 & 3 & 1 & 1 & 3 & 3 & 0 & 0 & Ca \\
Armadillo & 0 & 0 & 0 & 0 & 0 & 0 & 1 & 1 & Ci & Skunk & 3 & 3 & 1 & 1 & 3 & 3 & 0 & 0 & Ca \\
Pika & 2 & 1 & 0 & 0 & 2 & 2 & 1 & 1 & L & River Otter & 3 & 3 & 1 & 1 & 4 & 3 & 0 & 0 & Ca \\
Snowshoe Rabbit & 2 & 1 & 0 & 0 & 3 & 2 & 1 & 1 & L & Sea Otter & 3 & 2 & 1 & 1 & 3 & 3 & 0 & 0 & Ca \\
Beaver & 1 & 1 & 0 & 0 & 2 & 1 & 1 & 1 & R & Jaguar & 3 & 3 & 1 & 1 & 3 & 2 & 0 & 0 & Ca \\
Marmot & 1 & 1 & 0 & 0 & 2 & 1 & 1 & 1 & R & Ocelot & 3 & 3 & 1 & 1 & 3 & 2 & 0 & 0 & Ca \\
Groundhog & 1 & 1 & 0 & 0 & 2 & 1 & 1 & 1 & R & Cougar & 3 & 3 & 1 & 1 & 3 & 2 & 0 & 0 & Ca \\
Prairie Dog & 1 & 1 & 0 & 0 & 2 & 1 & 1 & 1 & R & Lynx & 3 & 3 & 1 & 1 & 3 & 2 & 0 & 0 & Ca \\
Ground Squirrel & 1 & 1 & 0 & 0 & 2 & 1 & 1 & 1 & R & Fur Seal & 3 & 2 & 1 & 1 & 4 & 4 & 0 & 0 & Ca \\
Chipmunk & 1 & 1 & 0 & 0 & 2 & 1 & 1 & 1 & R & Sea Lion & 3 & 2 & 1 & 1 & 4 & 4 & 0 & 0 & Ca \\
Gray Squirrel & 1 & 1 & 0 & 0 & 1 & 1 & 1 & 1 & R & Walrus & 1 & 0 & 1 & 1 & 3 & 3 & 0 & 0 & Ca \\
Fox Squirrel & 1 & 1 & 0 & 0 & 1 & 1 & 1 & 1 & R & Gray Seal & 3 & 2 & 1 & 1 & 3 & 3 & 0 & 0 & Ca \\
Pocket Gopher & 1 & 1 & 0 & 0 & 1 & 1 & 1 & 1 & R & Elephant Seal & 2 & 1 & 1 & 1 & 4 & 4 & 0 & 0 & Ca \\
Kangaroo Rat & 1 & 1 & 0 & 0 & 1 & 1 & 1 & 1 & R & Peccary & 2 & 3 & 1 & 1 & 3 & 3 & 1 & 1 & A \\
Pack Rat & 1 & 1 & 0 & 0 & 0 & 0 & 1 & 1 & R & Elk & 0 & 4 & 1 & 0 & 3 & 3 & 1 & 1 & A \\
Field Mouse & 1 & 1 & 0 & 0 & 0 & 0 & 1 & 1 & R & Deer & 0 & 4 & 0 & 0 & 3 & 3 & 1 & 1 & A \\
Muskrat & 1 & 1 & 0 & 0 & 0 & 0 & 1 & 1 & R & Moose & 0 & 4 & 0 & 0 & 3 & 3 & 1 & 1 & A \\
Black Rat & 1 & 1 & 0 & 0 & 0 & 0 & 1 & 1 & R & Reindeer & 0 & 4 & 1 & 0 & 3 & 3 & 1 & 1 & A \\
House Mouse & 1 & 1 & 0 & 0 & 0 & 0 & 1 & 1 & R & Antelope & 0 & 4 & 0 & 0 & 3 & 3 & 1 & 1 & A \\
Porcupine & 1 & 1 & 0 & 0 & 1 & 1 & 1 & 1 & R & Bison & 0 & 4 & 0 & 0 & 3 & 3 & 1 & 1 & A \\
Guinea Pig & 1 & 1 & 0 & 0 & 1 & 1 & 1 & 1 & R & Mountain Goat & 0 & 4 & 0 & 0 & 3 & 3 & 1 & 1 & A \\
Coyote & 1 & 3 & 1 & 1 & 4 & 4 & 1 & 1 & Ca & Muskox & 0 & 4 & 0 & 0 & 3 & 3 & 1 & 1 & A \\
Wolf & 3 & 3 & 1 & 1 & 4 & 4 & 0 & 1 & Ca & Mountain Sheep & 0 & 4 & 0 & 0 & 3 & 3 & 1 & 1 & A \\
\bottomrule
\end{tabular}%
}
\end{table}

\subsection{Mushroom Attributes: Nominal Structure and Edibility}

We next apply cGAP to the mushroom dataset from the UCI Machine Learning Repository~\citep{frank2010uci}, comprising 8,124 samples from the Agaricus and Lepiota families. Each sample is described by 22 nominal physical-attribute variables together with edibility information. This example tests whether cGAP can expose interpretable nominal structure in a large benchmark where class differences are strong but multivariable relationships remain visually complex.

\begin{figure}[h]
    \centering
    \includegraphics[width=0.9\textwidth]{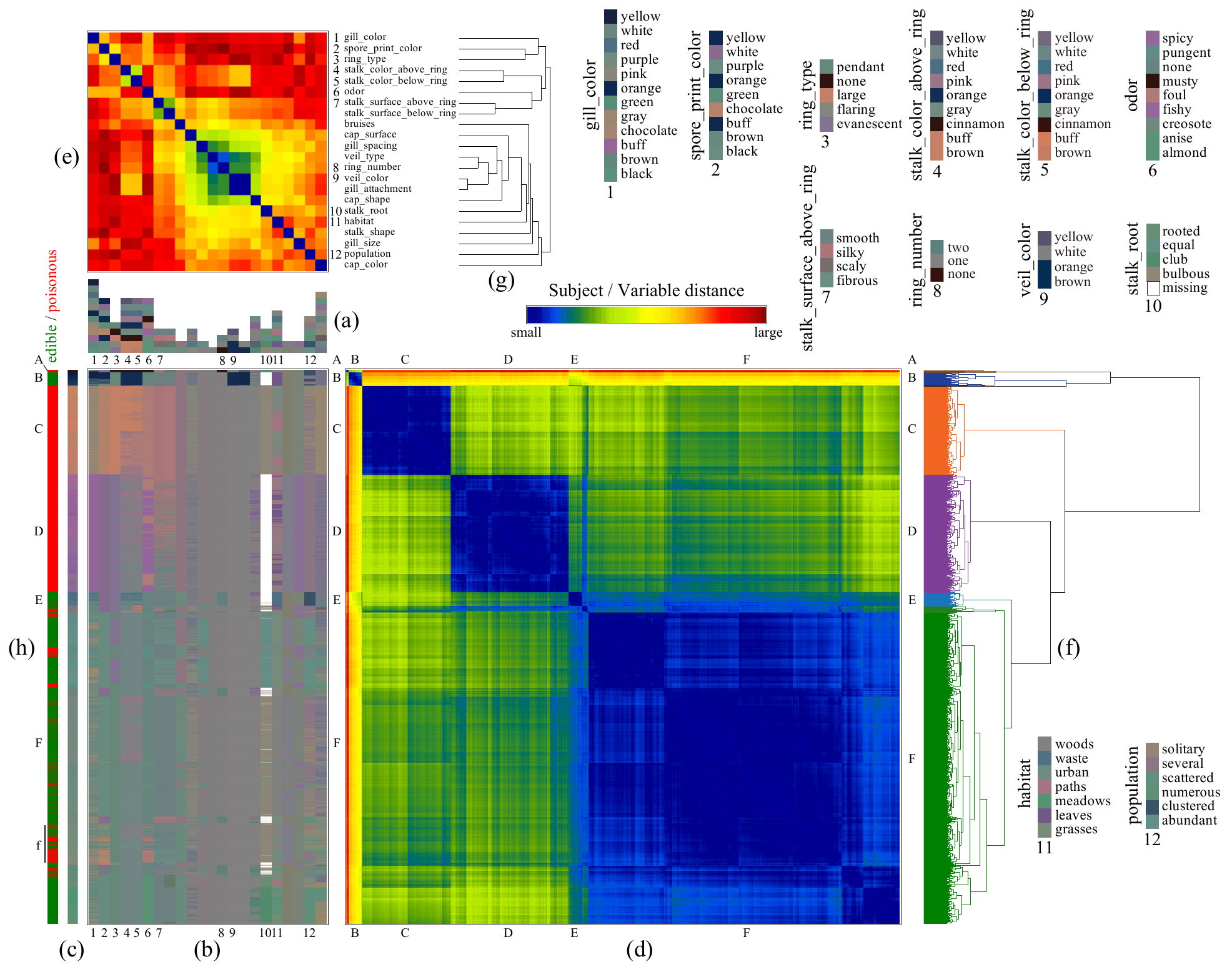} 
    \caption{cGAP analysis of the mushroom data (8,124 mushrooms $\times$ 22 nominal variables): (a) category color map, (b) sorted raw-data heatmap, (c) mushroom profile colors, (d) sorted Euclidean distance map for mushrooms, (e) sorted weighted distance map for variables, (f) HCT-R2E for (d), (g) HCT-R2E for (e), and (h) edibility labels for the samples.}
    \label{fig:figure6_mushroom}
\end{figure}

We use a three-dimensional HOMALS solution and HCT-R2E seriation~\citep{tien2008methods} to order both samples and variables. Figure \ref{fig:figure6_mushroom} presents the coordinated cGAP views. Because many local patterns repeat across mushrooms, the dataset is especially useful for distinguishing variables that contribute little separation from those that organize the major edible and poisonous subgroups.

Figure \ref{fig:figure6_mushroom} shows that only a subset of the 22 mushroom variables drives the major visual structure. The variable proximity view indicates that attributes such as cap surface, gill spacing, veil type, ring number, veil color, gill attachment, and cap shape contribute little discrimination, which is consistent with their relatively monotone appearance in the data matrix. By contrast, gill color, ring type, stalk color above ring, odor, and habitat show strong contrast and organize the main sample blocks, producing six visually coherent mushroom groups in the coordinated views.

\begin{figure}[htbp]
    \centering
    \includegraphics[width=0.65\textwidth]{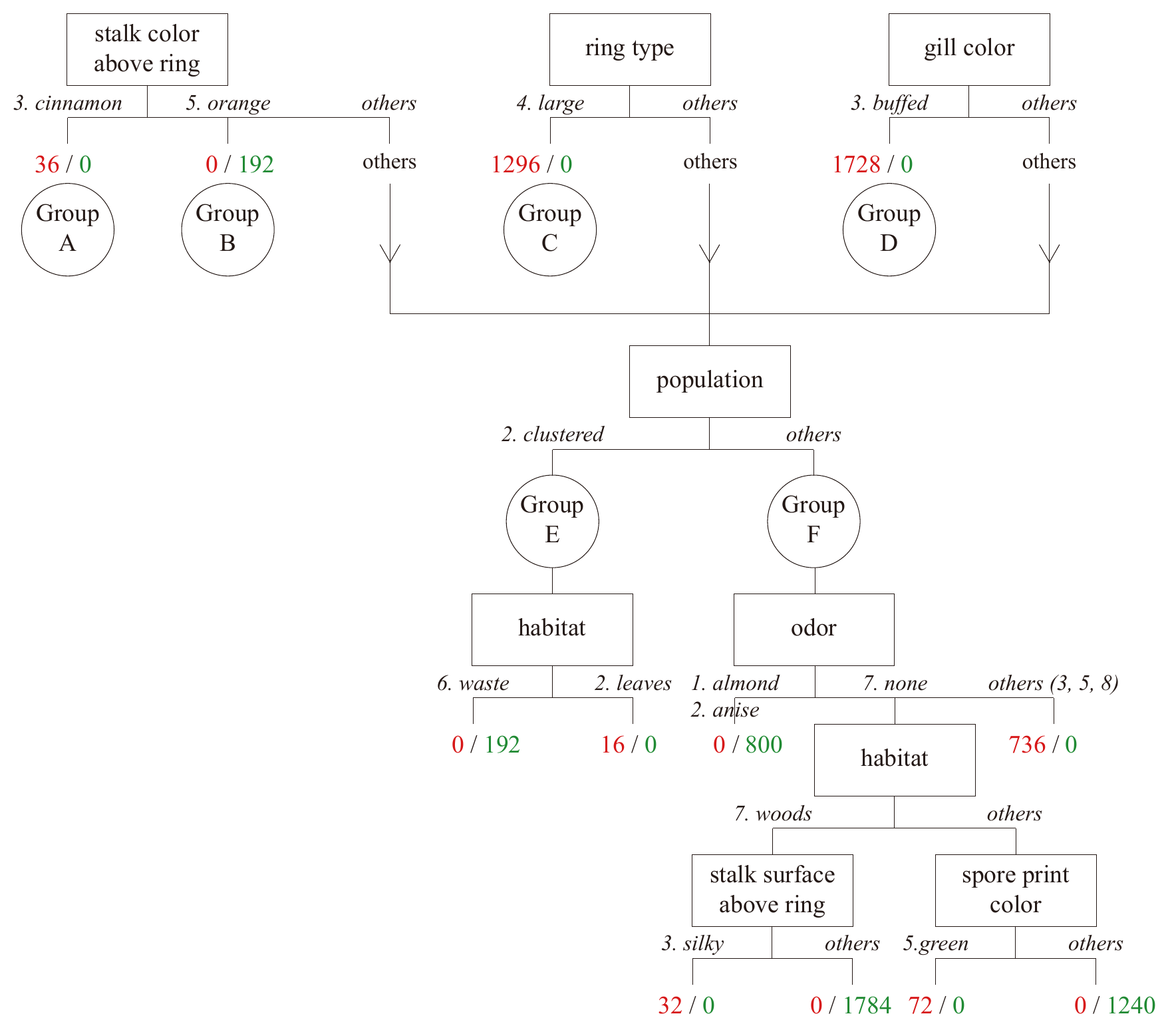} 
    \caption{One sequence of interpretable rules, derived from the cGAP views, for distinguishing poisonous from edible mushrooms using a reduced set of physical-attribute variables.}
    \label{fig:figure7_rules}
\end{figure}

These groups align closely with edibility patterns. Some clusters are almost entirely edible or poisonous, whereas mixed groups can be interpreted through a small number of variables, especially odor and habitat. The visualization also makes the missing-data structure transparent: missingness occurs only in the stalk-root variable and is closely associated with ring-type and ring-number patterns, suggesting a systematic rather than random mechanism. Figure \ref{fig:figure7_rules} summarizes one set of rules derived from this visual analysis. cGAP is not introduced here as a predictive classifier; rather, the mushroom example shows how the display can reduce a 22-variable nominal dataset to a smaller set of visually grounded and substantively interpretable decision cues.

This example also highlights a less obvious capability of cGAP: it distinguishes globally weak variables from locally decisive ones. Variables such as veil type or cap shape appear visually flat across most of the matrix, immediately identifying them as poor drivers of subgroup separation, whereas odor, habitat, gill color, and ring-related variables become decisive only within specific blocks of mushrooms. In that sense, the display does more than separate edible from poisonous samples; it reveals multiple local rule regimes for edibility rather than a single global decision boundary. The mixed clusters are especially informative because they show that cGAP can expose alternative pathways to similar class labels, such as habitat-based separation in one block and odor-based separation in another. The structured missingness in stalk-root strengthens this interpretation by showing that cGAP can simultaneously function as a pattern-discovery tool and a data-audit tool, revealing when apparently secondary variables are tied to systematic recording or biological substructure.

\subsection{COG Profiles: Large-Scale Binary Structure}

Our third example evaluates cGAP on large-scale binary data using the Clusters of Orthologous Genes (COG) database~\citep{tatusov1997genomic,tatusov2000cog,tatusov2001cog,galperin2015expanded,galperin2021cogupdate,galperin2024cogupdate}. We analyze 2,296 representative genomes (2,103 bacteria and 193 archaea) and 5,061 COG variables. Although the original database records paralog counts, we convert the table to a binary presence/absence matrix so that each entry indicates whether a genome contains a given COG. These phyletic profiles are widely used to study conserved functions, lineage-specific gene loss, and horizontal transfer~\citep{glazko2004detection,tzeng2009selection}.

\begin{figure}[t]
    \centering
    \includegraphics[width=0.75\textwidth]{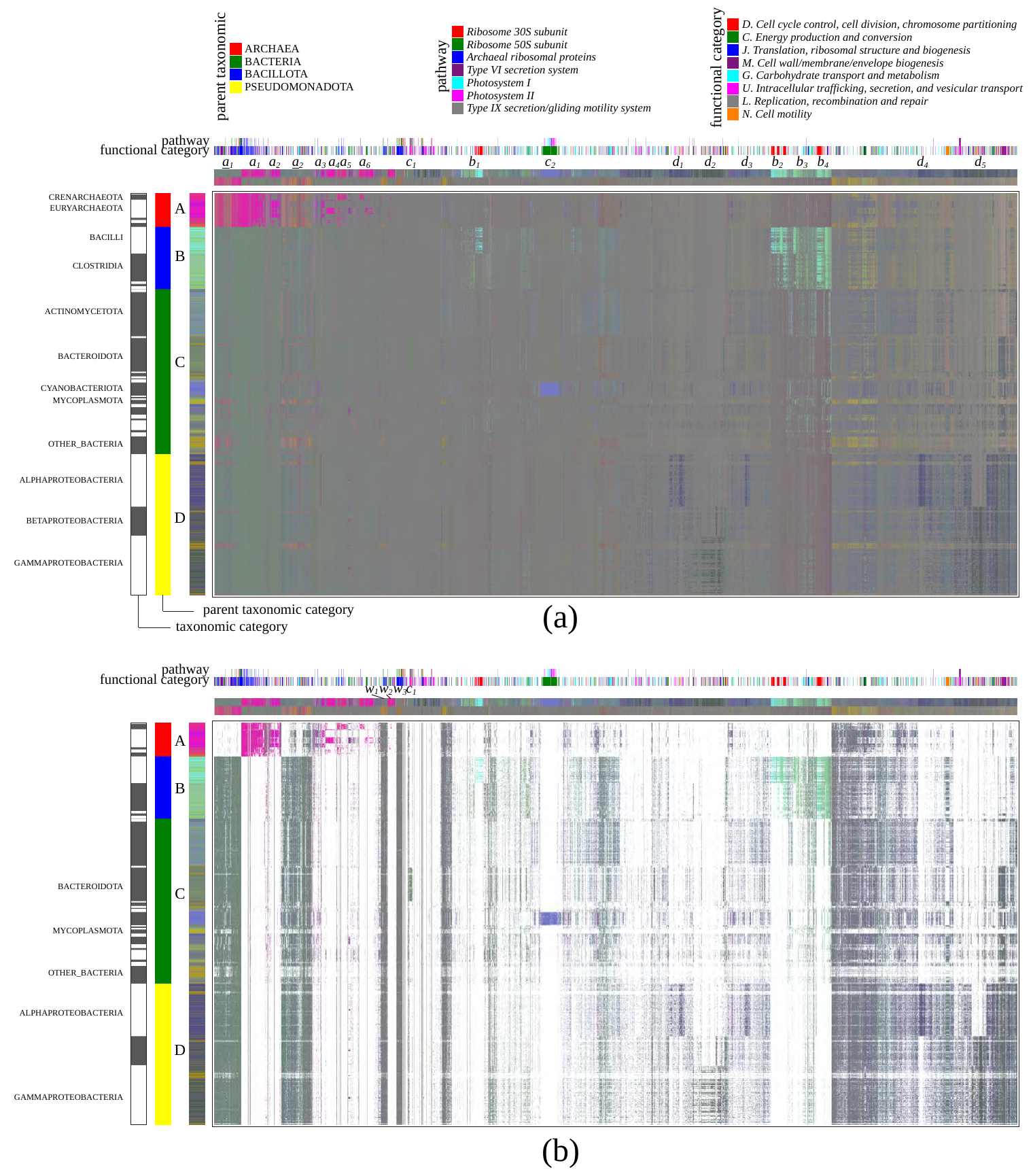} 
    \caption{cGAP analysis of the COG dataset (2,296 genomes $\times$ 5,061 variables): (a) raw-data heatmap with genomes kept in taxonomic order and COGs reordered by R1T-R2E seriation, and (b) a screening view that masks absent states and shows only present COGs. Only selected pathways and functional categories discussed in the text are labeled for readability.}
    \label{fig:figure8_cog}
\end{figure}

Within cGAP, the two binary states are treated as nominal categories and embedded jointly with genomes and COG variables. Figure \ref{fig:figure8_cog} emphasizes the matrix-scale view of the data: genomes retain their taxonomic ordering, whereas COGs are reordered by the R1T-R2E seriation algorithm. To keep the main figure readable at this scale, the primary panel focuses on the raw-data heatmap and the screening display; full-resolution proximity views are better suited to supplementary material in a journal presentation.

Figure \ref{fig:figure8_cog} demonstrates how cGAP scales to large binary data while preserving a readable matrix-centered view. ARCHAEA form a distinct block characterized by several archaeal COG groups, revealing both coarse separation from bacteria and finer internal subdivision, including multiple EURYARCHAEOTA clusters. The $a$ and $\underline{a}$ groups highlight an important advantage of cGAP for binary data: present and absent states can have inverse but closely related structure, so the method assigns them contrasting colors while keeping them close in the embedding. BACILLOTA and PSEUDOMONADOTA also show strong lineage-specific signatures, with the $b1-b4$ and $d1-d5$ groups differentiating major classes and connecting these blocks to secretion-, motility-, and membrane-related functions.

\begin{figure}[t]
    \centering
    \includegraphics[width=0.8\textwidth]{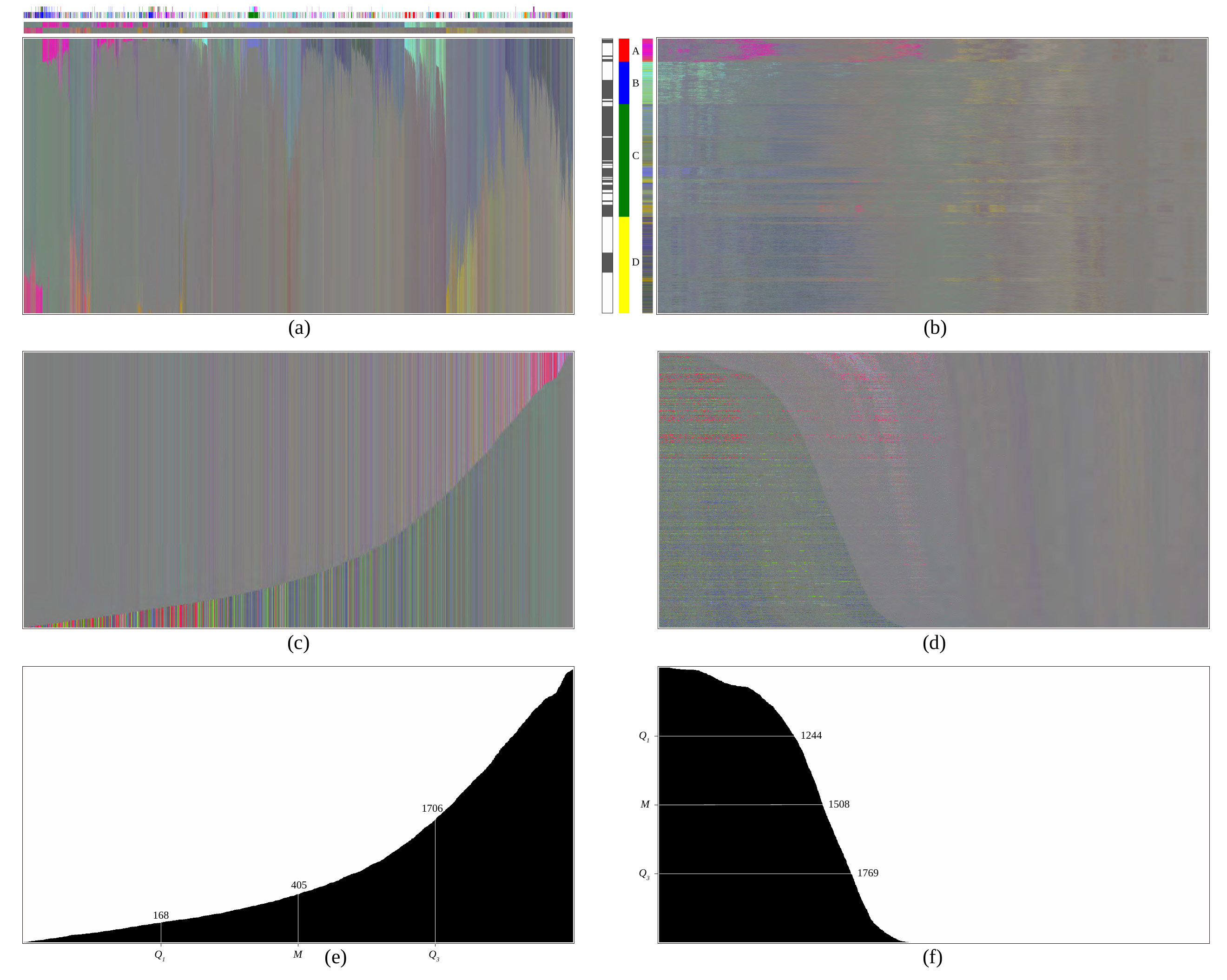} 
    \caption{Sediment displays for the COG dataset (2,296 genomes $\times$ 5,061 COGs): (a) COG-wise and (b) genome-wise plots following the arrangement in Figure \ref{fig:figure8_cog}(a), (c) and (d) the corresponding plots reordered by presence frequency, and (e) and (f) black-and-white presence/absence views of (c) and (d).}
    \label{fig:figure9_sediment}
\end{figure}

The screening view in Figure \ref{fig:figure8_cog}(b) reveals additional structure that is harder to see in the full matrix, including the association of BACTEROIDOTA with the $c1$ group, near-universal $w1$/$w3$ patterns, and genomes with near-total depletion of annotated COGs. This screening panel behaves like a presence-only overlay, but it is not equivalent to a conventional binary GAP based on Jaccard-type distances, because cGAP keeps complementary inverse patterns analytically connected through the HOMALS embedding. Figure \ref{fig:figure9_sediment} complements the matrix display with sediment plots summarizing marginal presence frequencies for COGs and genomes, showing that the method supports both local structural interpretation and global distributional screening in a large genomic application.

The COG application is particularly persuasive because it shows that cGAP can represent several layers of genomic structure in a single coordinated analysis. At the coarsest level, the raw-data heatmap preserves broad phylogenetic organization, separating archaeal and bacterial regimes. Within those regimes, the reordered COG blocks expose pathway- and function-level modules that refine the taxonomic picture, such as archaeal ribosomal signatures, Bacillota-specific cell-cycle patterns, cyanobacterial photosystem-related groups, and secretion- or motility-associated patterns within Pseudomonadota. The inverse $a$ and $\underline{a}$ relationships are especially important methodologically: they demonstrate that cGAP does not reduce binary structure to simple presence counts, but can keep complementary present and absent states close in the embedding while rendering them visually distinct. The screening view and sediment displays then add a multiscale perspective, making near-universal cores, lineage-specific systems, and genomes with unusually sparse annotations visible within the same analytic framework. Together, these features make the COG study a strong demonstration that cGAP remains interpretable even when the data are both high dimensional and biologically heterogeneous.

Across these three applications, cGAP shows consistent strengths: it preserves the raw categorical matrix, reveals subject- and variable-level structure through coordinated views, and remains interpretable across ordinal, nominal, and binary settings. More specifically, the applications show that cGAP can recover anatomical symmetry and functional gradients in ordinal data, distinguish globally weak variables from locally decisive ones in nominal data, and expose complementary present/absent modules together with multiscale phylogenetic structure in large binary data. Taken together, these examples suggest a practical presentation strategy for journal submission: the main paper can emphasize the integrated matrix displays and central findings, while large supporting proximity views can be provided as supplementary material when space is limited.

\section{Discussion and Conclusion}

This paper presents cGAP, a full-matrix visualization framework for high-dimensional categorical data that combines HOMALS embedding, color encoding, proximity matrices, and seriation in a coordinated display. The main contribution of cGAP is not simply to produce a low-dimensional view of categorical structure, but to preserve the original data matrix as a central analytic object while augmenting it with interpretable geometric and relational context. Across the examples in this paper, cGAP reveals clusters, outliers, complementary category patterns, and variable groupings in nominal, ordinal, and binary data.

The method is particularly useful when analysts need to move repeatedly between derived structure and raw observations. In contrast to standalone scatterplots or dimension-reduction displays, cGAP allows users to inspect every matrix entry while still benefiting from embedding-based color similarity and proximity-based organization. This makes the framework well suited to exploratory analysis tasks in which traceability, local pattern inspection, and multilevel comparison are all important. The barycentric traceability identity, projection-distortion decomposition, and ray-preserving contrast transform directly clarify why these displays can be interpreted in a principled way.

At the same time, cGAP has several limitations. First, the visual quality of the color encoding depends on how much categorical structure is retained by the chosen low-dimensional HOMALS solution. When the retained variation in three dimensions is limited, the color map may underrepresent some relationships even though the matrix display itself remains available. Second, RGB is used here as a practical display-native coordinate system rather than as a perceptually optimal color space. Although this design preserves the geometry-to-color mapping transparently, more perceptually calibrated and color-vision-deficiency-aware encodings would be valuable extensions. Third, very large datasets may require selective labeling, interactive exploration, or supplementary high-resolution views to fully expose subject and variable proximity structure in publication figures.

These limitations also suggest several directions for future work. One direction is to study alternative color spaces and constrained mappings that better balance interpretability, perceptual uniformity, and accessibility. Another is to extend cGAP with interactive interfaces that support zooming, brushing, linked highlighting, and user-controlled filtering for large categorical matrices. It would also be useful to investigate hybrid workflows in which higher-dimensional HOMALS solutions are used for proximity analysis while three-dimensional projections are reserved for display color construction.

In conclusion, cGAP provides an interpretable visualization environment for high-dimensional categorical data by linking the original matrix to embedding-based color structure and proximity-based organization. Rather than replacing existing multivariate methods, it complements them by offering a matrix-centered view that preserves data traceability while supporting exploratory discovery. We expect cGAP to be especially useful in domains where categorical structure is complex, high-dimensional, and scientifically important, including biomedical, genomic, and social-science applications.

\section*{Acknowledgment}
The authors thank the late Dr. Chih-Wen Ou-yang, a former postdoctoral fellow and research assistant, for invaluable discussions and insightful contributions to the development of the cGAP methodology. His dedication and intellectual input greatly shaped this work. This work was supported by the Ministry of Science and Technology, Taiwan, under the grant ``A comparison of matrix visualization and EDA of non-quantitative data'' (MOST 107-2118-M-001-004-MY2).

\section*{Software Availability}
The software described in this study is available for download at \url{https://gap.stat.sinica.edu.tw/software.html} and can also be accessed via an online version at \url{https://maokao.github.io/cGAPOnline/}.

\bibliographystyle{elsarticle-harv}
\bibliography{cGAP_Paper}

\begin{thebibliography}{27}
\expandafter\ifx\csname natexlab\endcsname\relax\def\natexlab#1{#1}\fi
\providecommand{\url}[1]{\texttt{#1}}
\providecommand{\href}[2]{#2}
\providecommand{\path}[1]{#1}
\providecommand{\DOIprefix}{doi:}
\providecommand{\ArXivprefix}{arXiv:}
\providecommand{\URLprefix}{URL: }
\providecommand{\Pubmedprefix}{pmid:}
\providecommand{\doi}[1]{\href{http://dx.doi.org/#1}{\path{#1}}}
\providecommand{\Pubmed}[1]{\href{pmid:#1}{\path{#1}}}
\providecommand{\bibinfo}[2]{#2}
\ifx\xfnm\relax \def\xfnm[#1]{\unskip,\space#1}\fi
\bibitem[{Benzecri(1973)}]{benzecri1973analyse}
\bibinfo{author}{Benzecri, J.P.}, \bibinfo{year}{1973}.
\newblock \bibinfo{title}{L'analyse des donnees. II. L'analyse des
  correspondances}.
\newblock \bibinfo{publisher}{Dunod}, \bibinfo{address}{Paris}.
\bibitem[{Chen(2002)}]{chen2002generalized}
\bibinfo{author}{Chen, C.H.}, \bibinfo{year}{2002}.
\newblock \bibinfo{title}{Generalized association plots: information
  visualization via iteratively generated correlation matrices}.
\newblock \bibinfo{journal}{Statistica Sinica} \bibinfo{volume}{12},
  \bibinfo{pages}{7--29}.
\bibitem[{Chen(2004)}]{chen2004matrix}
\bibinfo{author}{Chen, C.H.}, \bibinfo{year}{2004}.
\newblock \bibinfo{title}{Matrix visualization and information mining}, in:
  \bibinfo{booktitle}{COMPSTAT 2004 -- Proceedings in Computational
  Statistics}, pp. \bibinfo{pages}{85--100}.
\bibitem[{Frank and Asuncion(2010)}]{frank2010uci}
\bibinfo{author}{Frank, A.}, \bibinfo{author}{Asuncion, A.},
  \bibinfo{year}{2010}.
\newblock \bibinfo{title}{Uci machine learning repository}.
\newblock \URLprefix \url{http://archive.ics.uci.edu/ml}.
  \bibinfo{note}{irvine, CA: University of California, School of Information
  and Computer Science}.
\bibitem[{Friendly(1994)}]{friendly1994mosaic}
\bibinfo{author}{Friendly, M.}, \bibinfo{year}{1994}.
\newblock \bibinfo{title}{Mosaic displays for multi-way contingency tables}.
\newblock \bibinfo{journal}{Journal of the American Statistical Association}
  \bibinfo{volume}{89}, \bibinfo{pages}{190--200}.
\bibitem[{Friendly(1999)}]{friendly1999extending}
\bibinfo{author}{Friendly, M.}, \bibinfo{year}{1999}.
\newblock \bibinfo{title}{Extending mosaic displays: Marginal, conditional, and
  partial views of categorical data}.
\newblock \bibinfo{journal}{Journal of Computational and Graphical Statistics}
  \bibinfo{volume}{8}, \bibinfo{pages}{373--395}.
\bibitem[{Galperin et~al.(2015)Galperin, Makarova, Wolf and
  Koonin}]{galperin2015expanded}
\bibinfo{author}{Galperin, M.Y.}, \bibinfo{author}{Makarova, K.S.},
  \bibinfo{author}{Wolf, Y.I.}, \bibinfo{author}{Koonin, E.V.},
  \bibinfo{year}{2015}.
\newblock \bibinfo{title}{Expanded microbial genome coverage and improved
  protein family annotation in the {COG} database}.
\newblock \bibinfo{journal}{Nucleic Acids Research} \bibinfo{volume}{43},
  \bibinfo{pages}{D261--D269}.
\bibitem[{Galperin et~al.(2021)Galperin, Wolf, Makarova, Vera~Alvarez, Landsman
  and Koonin}]{galperin2021cogupdate}
\bibinfo{author}{Galperin, M.Y.}, \bibinfo{author}{Wolf, Y.I.},
  \bibinfo{author}{Makarova, K.S.}, \bibinfo{author}{Vera~Alvarez, R.},
  \bibinfo{author}{Landsman, D.}, \bibinfo{author}{Koonin, E.V.},
  \bibinfo{year}{2021}.
\newblock \bibinfo{title}{{COG} database update: focus on microbial diversity,
  model organisms and widespread pathogens}.
\newblock \bibinfo{journal}{Nucleic Acids Research} \bibinfo{volume}{49},
  \bibinfo{pages}{D274--D281}.
\bibitem[{Galperin et~al.(2024)}]{galperin2024cogupdate}
\bibinfo{author}{Galperin, M.Y.}, et~al., \bibinfo{year}{2024}.
\newblock \bibinfo{title}{{COG} database update 2024}.
\newblock \bibinfo{journal}{Nucleic Acids Research} \bibinfo{volume}{53},
  \bibinfo{pages}{D356--D363}.
\newblock \URLprefix \url{https://doi.org/10.1093/nar/gkae983},
  \DOIprefix\doi{10.1093/nar/gkae983}.
\bibitem[{Gifi(1990)}]{gifi1990nonlinear}
\bibinfo{author}{Gifi, A.}, \bibinfo{year}{1990}.
\newblock \bibinfo{title}{Nonlinear Multivariate Analysis}.
\newblock \bibinfo{publisher}{John Wiley \& Sons}.
\bibitem[{Glazko and Mushegian(2004)}]{glazko2004detection}
\bibinfo{author}{Glazko, G.V.}, \bibinfo{author}{Mushegian, A.R.},
  \bibinfo{year}{2004}.
\newblock \bibinfo{title}{Detection of evolutionarily stable fragments of
  cellular pathways by hierarchical clustering of phyletic patterns}.
\newblock \bibinfo{journal}{Genome Biology} \bibinfo{volume}{5},
  \bibinfo{pages}{R32}.
\newblock \URLprefix \url{https://doi.org/10.1186/2004-5-5-r32},
  \DOIprefix\doi{10.1186/2004-5-5-r32}.
\bibitem[{Greenacre(1984)}]{greenacre1984theory}
\bibinfo{author}{Greenacre, M.J.}, \bibinfo{year}{1984}.
\newblock \bibinfo{title}{Theory and Applications of Correspondence Analysis}.
\newblock \bibinfo{publisher}{Academic Press}, \bibinfo{address}{London}.
\bibitem[{Hartigan(1975)}]{hartigan1975clustering}
\bibinfo{author}{Hartigan, J.}, \bibinfo{year}{1975}.
\newblock \bibinfo{title}{Clustering Algorithms}.
\newblock \bibinfo{publisher}{Wiley}, \bibinfo{address}{New York, NY}.
\bibitem[{Hartigan and Kleiner(1981)}]{hartigan1981mosaics}
\bibinfo{author}{Hartigan, J.A.}, \bibinfo{author}{Kleiner, B.},
  \bibinfo{year}{1981}.
\newblock \bibinfo{title}{Mosaics for contingency tables}, in:
  \bibinfo{editor}{Eddy, W.F.} (Ed.), \bibinfo{booktitle}{Computer Science and
  Statistics: Proceedings of the 13th Symposium on the Interface}.
  \bibinfo{publisher}{Springer-Verlag}, \bibinfo{address}{New York}.
\bibitem[{Kumasaka and Shibata(2008)}]{kumasaka2008high}
\bibinfo{author}{Kumasaka, N.}, \bibinfo{author}{Shibata, R.},
  \bibinfo{year}{2008}.
\newblock \bibinfo{title}{High-dimensional data visualisation: The textile
  plot}.
\newblock \bibinfo{journal}{Computational statistics \& data analysis}
  \bibinfo{volume}{52}, \bibinfo{pages}{3616--3644}.
\bibitem[{de~Leeuw et~al.(1967)de~Leeuw, Young and
  Takane}]{deleeuw1967additive}
\bibinfo{author}{de~Leeuw, J.}, \bibinfo{author}{Young, F.W.},
  \bibinfo{author}{Takane, Y.}, \bibinfo{year}{1967}.
\newblock \bibinfo{title}{Additive structure in qualitative data: An
  alternating least squares method with optimal scaling features}.
\newblock \bibinfo{journal}{Psychometrika} \bibinfo{volume}{41},
  \bibinfo{pages}{471--503}.
\bibitem[{Michailidis and de~Leeuw(1998)}]{michailidis1998gifi}
\bibinfo{author}{Michailidis, G.}, \bibinfo{author}{de~Leeuw, J.},
  \bibinfo{year}{1998}.
\newblock \bibinfo{title}{The {Gifi} system of descriptive multivariate
  analysis}.
\newblock \bibinfo{journal}{Statistical Science} \bibinfo{volume}{13},
  \bibinfo{pages}{307--336}.
\bibitem[{Minnotte and West(1998)}]{minnotte1998data}
\bibinfo{author}{Minnotte, M.}, \bibinfo{author}{West, R.W.},
  \bibinfo{year}{1998}.
\newblock \bibinfo{title}{The {Data Image}: A tool for exploring high
  dimensional data sets}, in: \bibinfo{booktitle}{Proceedings of the Section on
  Statistical Graphics, American Statistical Association},
  \bibinfo{address}{Alexandria, Virginia}.
\bibitem[{Nishisato(1984)}]{nishisato1984dual}
\bibinfo{author}{Nishisato, S.}, \bibinfo{year}{1984}.
\newblock \bibinfo{title}{Dual scaling of reciprocal medians}, in:
  \bibinfo{booktitle}{Proceedings of the 32nd Scientific Conference of the
  Italian Statistical Society}, \bibinfo{publisher}{Societa Italiana di
  Statistica}, \bibinfo{address}{Sorrento, Italy}. pp.
  \bibinfo{pages}{141--147}.
\bibitem[{Nishisato(2007)}]{nishisato2007multidimensional}
\bibinfo{author}{Nishisato, S.}, \bibinfo{year}{2007}.
\newblock \bibinfo{title}{Multidimensional Nonlinear Descriptive Analysis}.
\newblock \bibinfo{publisher}{Chapman \& Hall}.
\bibitem[{Tatusov et~al.(2000)Tatusov, Galperin, Natale and
  Koonin}]{tatusov2000cog}
\bibinfo{author}{Tatusov, R.L.}, \bibinfo{author}{Galperin, M.Y.},
  \bibinfo{author}{Natale, D.A.}, \bibinfo{author}{Koonin, E.V.},
  \bibinfo{year}{2000}.
\newblock \bibinfo{title}{The {COG} database: a tool for genome-scale analysis
  of protein functions and evolution}.
\newblock \bibinfo{journal}{Nucleic Acids Research} \bibinfo{volume}{28},
  \bibinfo{pages}{33--36}.
\bibitem[{Tatusov et~al.(1997)Tatusov, Koonin and Lipman}]{tatusov1997genomic}
\bibinfo{author}{Tatusov, R.L.}, \bibinfo{author}{Koonin, E.V.},
  \bibinfo{author}{Lipman, D.J.}, \bibinfo{year}{1997}.
\newblock \bibinfo{title}{A genomic perspective on protein families}.
\newblock \bibinfo{journal}{Science} \bibinfo{volume}{278},
  \bibinfo{pages}{631--637}.
\bibitem[{Tatusov et~al.(2001)Tatusov, Natale, Garkavtsev, Tatusova,
  Shankavaram, Rao, Kiryutin, Galperin, Fedorova and Koonin}]{tatusov2001cog}
\bibinfo{author}{Tatusov, R.L.}, \bibinfo{author}{Natale, D.A.},
  \bibinfo{author}{Garkavtsev, I.V.}, \bibinfo{author}{Tatusova, T.A.},
  \bibinfo{author}{Shankavaram, U.T.}, \bibinfo{author}{Rao, B.S.},
  \bibinfo{author}{Kiryutin, B.}, \bibinfo{author}{Galperin, M.Y.},
  \bibinfo{author}{Fedorova, N.D.}, \bibinfo{author}{Koonin, E.V.},
  \bibinfo{year}{2001}.
\newblock \bibinfo{title}{The {COG} database: new developments in phylogenetic
  classification of proteins from complete genomes}.
\newblock \bibinfo{journal}{Nucleic Acids Research} \bibinfo{volume}{29},
  \bibinfo{pages}{22--28}.
\bibitem[{Tien et~al.(2008)}]{tien2008methods}
\bibinfo{author}{Tien, Y.J.}, et~al., \bibinfo{year}{2008}.
\newblock \bibinfo{title}{Methods for simultaneously identifying coherent local
  clusters with smooth global patterns in gene expression profiles}.
\newblock \bibinfo{journal}{BMC Bioinformatics} \bibinfo{volume}{9},
  \bibinfo{pages}{1--16}.
\newblock \bibinfo{note}{Article 155}.
\bibitem[{Tzeng et~al.(2009)Tzeng, Wu and Chen}]{tzeng2009selection}
\bibinfo{author}{Tzeng, S.L.}, \bibinfo{author}{Wu, H.M.},
  \bibinfo{author}{Chen, C.H.}, \bibinfo{year}{2009}.
\newblock \bibinfo{title}{Selection of proximity measures for matrix
  visualization of binary data}, in: \bibinfo{booktitle}{Proceedings of the
  2009 2nd International Conference on BioMedical Engineering and Informatics
  ({BMEI} 2009)}, pp. \bibinfo{pages}{1932--1940}.
\bibitem[{Wegman(1990)}]{wegman1990hyperdimensional}
\bibinfo{author}{Wegman, E.J.}, \bibinfo{year}{1990}.
\newblock \bibinfo{title}{Hyperdimensional data analysis using parallel
  coordinates}.
\newblock \bibinfo{journal}{Journal of the American Statistical Association}
  \bibinfo{volume}{85}, \bibinfo{pages}{664--675}.
\bibitem[{Wu et~al.(2010)Wu, Tien and Chen}]{wu2010gap}
\bibinfo{author}{Wu, H.M.}, \bibinfo{author}{Tien, Y.J.},
  \bibinfo{author}{Chen, C.H.}, \bibinfo{year}{2010}.
\newblock \bibinfo{title}{{GAP}: A graphical environment for matrix
  visualization and cluster analysis}.
\newblock \bibinfo{journal}{Computational Statistics and Data Analysis}
  \bibinfo{volume}{54}, \bibinfo{pages}{767--778}.

\end{thebibliography}

\end{document}